\documentclass[Afour,sageh,times]{sagej}

\usepackage{moreverb,url}
\usepackage[colorlinks,bookmarksopen,bookmarksnumbered,citecolor=red,urlcolor=red]{hyperref}
\usepackage{subcaption}
\usepackage{bm}
\usepackage{algorithmic}
\usepackage[linesnumbered,lined,ruled]{algorithm2e}
\newcommand\BibTeX{{\rmfamily B\kern-.05em \textsc{i\kern-.025em b}\kern-.08em
T\kern-.1667em\lower.7ex\hbox{E}\kern-.125emX}}

\def\volumeyear{2016}
\begin{document}
 
\runninghead{S. Sidhik, M. Sridharan, D. Ruiken}

\title{An Adaptive Framework for Reliable Trajectory Following in Changing-Contact Robot Manipulation Tasks}

\author{Saif Sidhik\affilnum{1}, Mohan Sridharan\affilnum{1} and Dirk Ruiken\affilnum{2}}

\affiliation{\affilnum{1}Intelligent Robotics Lab, University of Birmingham, UK\\
\affilnum{2}Honda Research Institute Europe GmbH, Germany}

\corrauth{Saif Sidhik,
Intelligent Robotics Lab,
School of Computer Science,
University of Birmingham,
Birmingham, UK.}

\email{sxs1412@bham.ac.uk}

\begin{abstract}
We describe a framework for changing-contact robot manipulation tasks that require the robot to make and break contacts with objects and surfaces. The discontinuous interaction dynamics of such tasks make it difficult to construct and use a single dynamics model or control strategy, and the highly non-linear nature of the dynamics during contact changes can be damaging to the robot and the objects. We present an adaptive control framework that enables the robot to incrementally learn to predict contact changes in a changing contact task, learn the interaction dynamics of the piece-wise continuous system, and provide smooth and accurate trajectory tracking using a task-space variable impedance controller. We experimentally compare the performance of our framework against that of representative control methods to establish that the adaptive control and incremental learning components of our framework are needed to achieve smooth control in the presence of discontinuous dynamics in changing-contact robot manipulation tasks.
\end{abstract}


\maketitle

\section{Introduction}
\label{sec:intro}
\begin{figure}[tb]
  \begin{center}
   \begin{subfigure}{\columnwidth}
      \includegraphics[width=\linewidth]{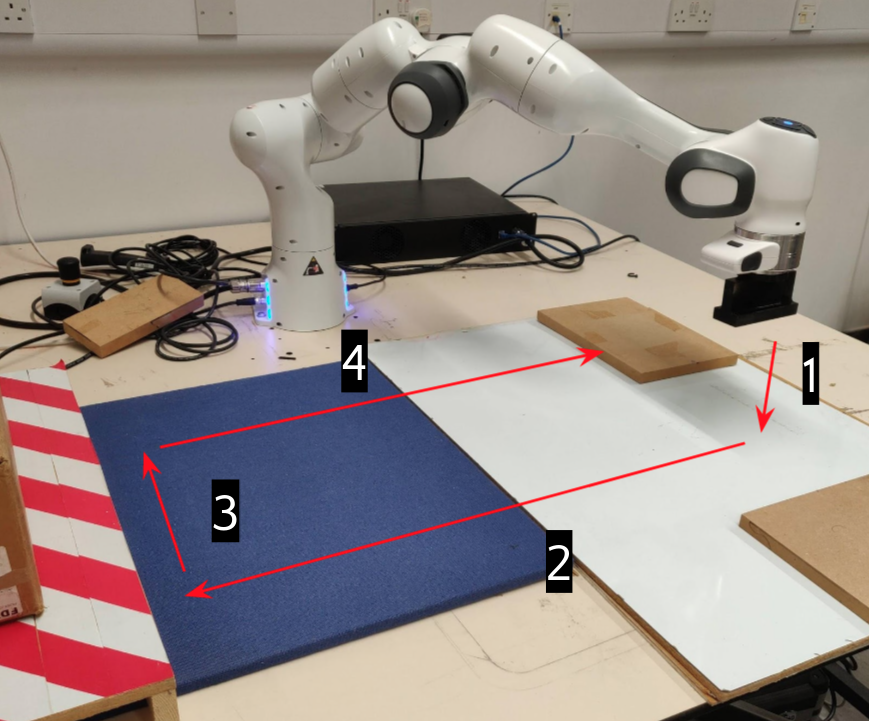}
      \caption{Changing-contact task.}
      \label{fig:exp-impactless_task}
   \end{subfigure}

   \begin{subfigure}{0.49\columnwidth}
      \includegraphics[width=0.95\linewidth]{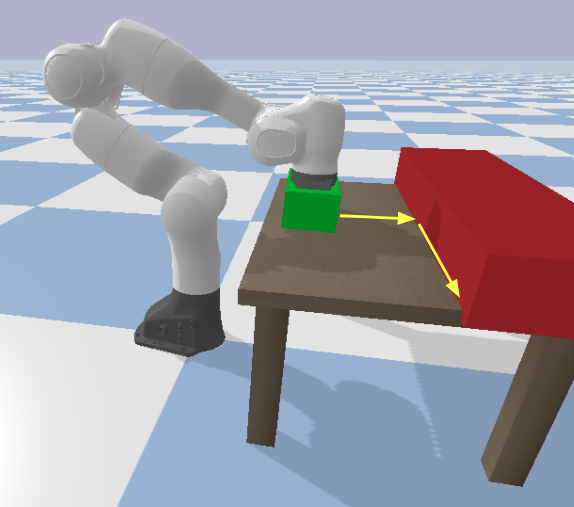}
      \caption{3D simulation setup.}
      \label{fig:pb-shahbaz-task}
   \end{subfigure}
\hfill
   \begin{subfigure}{0.49\columnwidth}
      \includegraphics[width=\linewidth]{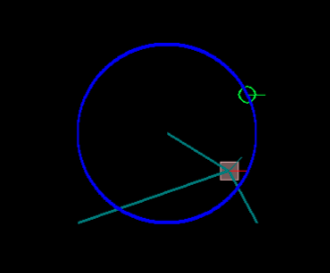}
      \caption{2D test scenario.}
      \label{fig:exp-ctrl-compare-aic-lag-screenshot}
   \end{subfigure}
  \end{center}
  \caption{(a) Changing-contact task where the robot experiences discontinuities in dynamics due to different surface friction in the middle of motions ``2'' and ``4'', and due to collisions at the end of ``1'', ``2'' and ``4''; (b) Simulated environment used in experiments in Section \ref{sec:exp-hyb-sys}; robot approaches table from top, slides end-effector (green) along table until it collides with wall, and slides along wall and table; (c) 2D \textit{multi-spring environment} with the robot lagging behind target using AIC (used in Sections \ref{sec:need_for_inc_models} and \ref{sec:ctrl-compare}): red block is robot end-effector, green lines are springs attached to end-effector.}
  \label{fig:intro-examples}
\end{figure}

Consider the robot manipulator in Figure~\ref{fig:exp-impactless_task} that has to move its end-effector along a motion pattern that requires it to make and break contact with objects and surfaces. This task's dynamics, i.e., the relationship between the forces acting on the robot and the resultant accelerations, vary markedly before and after contact (e.g. after motion `1' in the figure). They also vary based on other factors such as type of contact, surface friction (e.g. in the middle of motions `2' and `4'), and applied force. Many industrial assembly tasks, e.g., peg insertion, screwing and stacking, and many human manipulation tasks are instances of such `\textit{changing-contact}' tasks. The interaction dynamics of the robot performing these tasks are discontinuous when a contact is made or broken and continuous elsewhere, making it difficult to construct a single dynamics model. It is possible to construct a \emph{hybrid} model with separate continuous dynamics and distinct control laws within each of a number of discrete \emph{dynamic modes}; the overall dynamics are then \emph{piece-wise continuous}, with the robot transitioning between modes as needed~\citep{kroemer2019review}. Even then, the highly non-linear nature of such interactions results in large discontinuities in the overall motion dynamics of the robot at the regions of transition, in the form of high forces, jerk, and vibrations. These sudden and uncontrolled system disturbances during contact change instances such as collisions can damage the robot and/or objects involved. Therefore, the robot also has to learn to handle these discontinuities such that the overall motion is smooth and safe.

An ideal and efficient control framework for executing a provided manipulation task plan should therefore be able to follow the provided plan accurately while ensuring that: (i) it uses low stiffness parameters whenever possible so as to be compliant and expend less energy; (ii) it can handle continuously changing interaction dynamics that arise when the robot is in continuous contact with an object/environment as part of the task; (iii) it can handle the piecewise continuous nature of interaction dynamics that occur due to discrete changes in the interacting environment or due to making/breaking of contacts; (iv) it can quickly learn to deal with previously unseen environments (interaction dynamics) and use an appropriate adaptive controller for navigating it; (v) it can learn to anticipate collisions and contact changes in the task, and be able to learn and switch to a control strategy that produce reduced impact forces and vibrations at these transitions, (vi) the robot should require as few trials as possible to learn to perform the task efficiently (where `efficiency' is defined in terms of trajectory-tracking accuracy, controller stiffness, delays in task-completion, low impact forces, and overall smoothness of motion).


This paper describes an online adaptive framework for smooth control of changing-contact robot manipulation tasks that tries to satisfy these conditions. It unifies our work on a variable impedance controller for continuous contact tasks~\citep{humanoids_varimp}, an extension for piecewise-continuous dynamics due to contact changes~\citep{sidhik_learning}, and a recent study which explored a contact-change-handling module for dealing with the discontinuities in dynamics during contact changes~\citep{sidhik:iros21}. We advocate the need for a hybrid model with one or more  dynamics modes, each modeled for a contact mode with a predictive (forward) dynamics model of sensor measurements, a control law, and a relevance condition. The choice of representation for learning the hybrid models enables efficient adaptation of the dynamics model of each mode and makes the identification of modes independent of the motion direction and magnitude of applied forces. Our controller within a mode is inspired by research in human motor control, which indicates that people learn to adapt arm stiffness to any new task by building and using internal models of task dynamics with generic (domain/task-independent) and specific (domain/task-dependent) representations to predict hand or object configurations and the forces during task execution~\citep{kawato1999internal}. 

For sensor measurements and system feedback, we rely purely on the robot's end-effector poses/velocities and force-torque estimates at the end-effector of the robot. Our framework is capable of using external sensors such as cameras for providing initial estimates of the contact modes and/or contact locations, but assumes these are not always available due to their unreliability in the context of changing-contact manipulation tasks. The motion pattern, predictors, and control laws were defined in Cartesian space to simplify the learning and control problems, as well as to make the framework more generalisable and independent of the robot platform. The contribution of this paper is a study of the unified framework that combines the following three main components:
\begin{enumerate}

\item An adaptive variable impedance control strategy to be used within a continuous contact mode, with the capability to automatically and incrementally adapt the forward model of the mode's dynamics, which is used to revise the gain (i.e., stiffness) parameters of the control law for accurate and compliant motion in that mode.

\item A hierarchical model learning approach to model piecewise-continuous dynamics of changing-contact manipulation tasks without prior knowledge of all its modes or the order in which they appear. The approach also includes a mode-detection model that automatically identifies the modes of any given task, and transitions to existing or new modes during task execution.

\item A simple and efficient contact-change-handling module that can predict and incrementally improve the estimates of contact locations, use the estimates to minimize time spent in the transition phase, and achieve smooth motion dynamics during mode transitions by adapting controller parameters automatically.
\end{enumerate}
The framework is tested in changing-contact manipulation setups such as those shown in Figure~\ref{fig:intro-examples}. Components and features of the framework are evaluated against existing baselines mainly in contained simulated setups for providing noiseless and unbiased results. The adaptive variable impedance control component for continuous dynamics (base controller) is experimentally compared with other adaptive control strategies from literature in a custom-built simulated world with different types of continuously changing environments. Then, the need for an incremental hybrid learning \& control strategy for handling discontinuities in dynamics is demonstrated using as baseline a fully offline long-term prediction framework, which is then incrementally modified to show the advantages of certain representational choices and formulation of our hybrid framework. 
We first review related work in Section~\ref{sec:related-work}, describe our framework in Section~\ref{sec:framework}, discuss experimental results in Section~\ref{sec:expres}, and conclude in Section~\ref{sec:conclusion}.

\section{Related work} 
\label{sec:related-work}

A robot manipulator typically interacts with the environment through contacts. This can be discrete such as when making/breaking contact (e.g., assembly tasks) or continuous (e.g., polishing, deburring). This section will review some works from literature that attempted to learn such piecewise continuous interactions and developed controllers for handling non-linear systems.

\subsection{Forward models in manipulation}\label{sec:lr_fm}

Since it is difficult to provide the robot the comprehensive knowledge of task and domain required to perform a desired manipulation task, the robot is often enabled to learn from experience, e.g., create a predictive (\emph{forward}) model of action effects and use it to select actions~\citep{beetz2010generality}. Studies in human motor control also indicate the creation and use of forward-models for different control tasks~\citep{flanagan2003prediction}.


Forward models broadly fall into three categories (i) analytic models - which are typically modeled mathematically using Newtonian physics; (ii) learned models - built from data using techniques from machine learning; and (iii) hybrid models - a combination of learned and analytic models. Analytic models are informed by the knowledge of mechanics to make predictions about robot and object motions \citep{fan2017real, liu2009dextrous}, by making assumptions such as quasi-static mechanics, zero slippage and point contacts \citep{chatterjee1999realism,fazeli2020fundamental}. Hence analytic methods often suffer from the problem of inaccurate prediction.  They also often require an explicit representation of its intrinsic parameters, such as friction, mass, mass distribution, and coefficients of restitution, which are not trivial to estimate \citep{kopicki2017learning}. 

The second approach is to use techniques from machine learning to build forward models of the system. These techniques learn an action-effect correlation either using data obtained from expert demonstrations \citep{kronander2014learning, huang2016modular} or from self experience using trials \citep{levine2013guided, kupcsik2013data}. Many methods have been developed to address the learning and control problems in robot manipulation~\citep{kroemer2019review}, especially methods based on reinforcement learning (RL)~\citep{stulp2012reinforcement} and those combining deep networks and RL for learning flexible behaviors from complex data~\citep{andrychowicz2018learning,hausman2018learning}.
Recently, several attempts have been made to use deep neural networks (DNNs) for end-to-end learning of changing-contact manipulation tasks which bypasses learning a separate policy for modeling dynamics \citep{nagabandi2020deep,ajay2019combining}. 
Although these methods reduce the need for domain models and prior knowledge, they require large labeled datasets, pose high-dimensional optimization  problems, and tend to select the smoothest interpolation of the training data, which conflicts with the discontinuous impact dynamics of changing-contact manipulation. 

The third approach is to use a combination of both analytical and learning approaches discussed above. These forward models tend to learn the difference between analytical models and the true dynamics of the interactions, capturing the improvements to the analytical model needed to match the observed environment (\citet{gandhi2017pseudospectral}). Despite its advantage of requiring less training data, these strategies still rely on several restrictive assumptions regarding the type of contacts, friction models, object dynamics, etc., require significant knowledge about the mathematical models to be used, and also require at least a few trials and examples of the real-world possible scenarios to be modeled. Even with sim-to-real strategies that were developed to further reduce the training on real robots, aspects such as the dynamics of rigid bodies with friction are too complicated to be modeled in a real-time dynamics simulator~\citep{johnson2016hybrid}, and often require several hundreds of trials in the real world for fine-tuning the models for transferring to a physical robot \citep{ajay2019combining}. 

The piece-wise continuous nature of changing contact tasks has been used by some methods to build hybrid or hierarchical models, with one level identifying the current ``mode" while the other level models the dynamics of that mode~\citep{bucsoniu2018reinforcement}. Planning methods for manipulation often consider the discontinuities but assume prior knowledge of the models and modes, and often require many synthetically-generated training samples~\citep{toussaint2018differentiable}. Our framework, on the other hand, incrementally learns the dynamics of the robot's interaction with its environment as a hierarchical model without prior knowledge (other than the target trajectory) of the dynamics.

\subsection{Controllers for continuous contact tasks}

In most practical control systems including robot manipulators, the model of the ``plant" (robot) is nonlinear or unknown. In case of manipulators, the model sometimes can be unmodelled or difficult to model due to interactions with other objects, unknown tools/objects at the end-effector, etc. This raises the need for control schemes that are capable of handling system uncertainties, or are adaptive and can change according to the system's response and/or observations (measurements). 

Adaptive control is class of controllers that are used for systems whose parameters can change, or are unknown beforehand, using feedback from the system. Common strategies include model-reference adaptive controllers (MRAC), self-tuning regulators, and gain scheduling. MRAC methods compute control signals that forces the system's behavior to match that of a reference model~\citep{zhang2017review}. They often employ approximations of the system's parameters and revise them using prior/observed data, but an incorrect reference model can result in instabilities. 

Self-tuning regulators typically model the plant as a linear time-varying system, and adapt or estimate the controller parameters online~\citep{pezzato2020novel}. Although they can converge to an optimal controller under certain conditions and are a good option for systems whose dynamics vary in a fixed pattern across all trials, they often struggle to adapt to the changing objectives in robot manipulation. 
In repetitive tasks, however, the last estimates of the parameters from the previous run can be used as the initial estimates, which can then be improved in subsequent trials. Thus, the controller is ``trainable" in repetitive tasks, resulting in a more accurate performance for systems. Under such situation, a controller is able to be designed or trained for every single anticipated operating point. This type of control adaptation falls under the next category of adaptive control, called gain-scheduling.

Gain-scheduling methods are a popular option in robotics due to the increasing use of imitation learning and RL methods. These methods learn a time-indexed sequence of control parameters by repeating the task to update gains~\citep{kramberger2018passivity}, or by learning gain profiles from training data~\citep{lee2015learning}). 
Many of the variable impedance strategies using reinforcement learning or learning-from-demonstrations can be considered to be in this category \citep{abu2018force,calinon2010learning,rozo2016learning}.
Variable impedance controllers in general provide a time-varying impedance profile for the robot controller during task execution. 
The impedance could change as a function of time, robot state, or any observation made by the robot. 
Since these methods often need large training datasets or comprehensive domain knowledge, more recent methods have the agent repeat a task till some measure of good performance is achieved~\citep{gams2014coupling,kramberger2018passivity}. Conventional methods try to reduce trajectory tracking error as well as improve rejection of \textit{periodic} disturbances. This typically involves learning a corrective term for the control law that linearly depends on tracking error, measured disturbance, and/or time. 
  

The adaptive variable impedance control (AVIC) we developed for navigating each contact mode is (at its core) similar to the self-tuning regulators, but it builds on our prior work in control for continuous-contact tasks~\citep{humanoids_varimp} and piecewise-continuous interaction dynamics~\citep{sidhik_learning}; it uses the prediction error of an incrementally learned forward model to guide the online adaptation of the control parameters. 
\subsection{Control of hybrid systems}\label{sec:lr-hyb_sys_ctrl}

Several control methods have been proposed in the control theory literature for specific classes of hybrid systems. For instance, a wide range of literature can be found for switched systems where stabilizing controllers are developed using Lyapunov arguments \citep{johansson2003piecewise,de2009survey}. 
Optimal control theory has been used for developing control approaches for hybrid systems in the context of manufacturing \citep{de2009survey}. Recently, model predictive control (MPC) has also been extended to some classes of hybrid systems \citep{de2009survey}.

Benefits of incorporating modes or phases in the design of controllers for manipulation is evident in grasping tasks (e.g. ~\citep{romano2011human}) where the controller has distinct objectives and behavior requirements for phases such as `approach', `grasp' and `release'. Different strategies for sequencing motion primitives have also been used to solve manipulation tasks, but they assume the existence of a library of modes or motion primitives, or alternatively segment a sequence of primitives from human  demonstrations~\citep{niekum2013incremental}. These methods do not consider the interaction dynamics or try to reduce effects of impact. Therefore, this type of modeling makes the learned policy dependent on the environment, movements and the sequence of modes. 

Motivated from the effectiveness and flaws from the works discussed in the previous sections, our framework uses independent task-space adaptive variable impedance controllers (AVIC) for each identified mode in the task. This makes the framework more generalisable and adaptable to new environments.

\subsection{Modeling impacts \& contact changes} \label{sec:lr-handling_contacts}

Collisions and impact dynamics introduce critical challenges to planning, modeling and control of robots in applications such as locomotion \citep{wieber2016modeling} and manipulation \citep{kemp2007challenges}. Even a single collision is a complex interaction where object interpenetration is prevented by
material deformation, and which often occurs on a scale far below the resolution of practical sensors \citep{halm2021set}. Capturing these processes accurately requires an impractically precise set of knowledge regarding the materials, geometries, and initial conditions, on top of the complications in predicting dynamics using these information \citep{chatterjee1997rigid}. To avoid these challenges, most robotics approaches make several coarse approximations of contact dynamics such as the rigid-body assumption to make the problem more tractable (see \citep{brogliato2019nonsmooth} for background).  When impacts occur, rigid-body models approximate the event as an instantaneous change in velocity due to an impulsive force. Even so, seemingly minor changes in the mathematical models can result in significantly different predictions from identical initial conditions, and in many cases, they are unable to capture real-world behaviors with any available model \citep{fazeli2020fundamental, stoianovici1996critical}.

Unrealistic contact models is a primary reason for the gap between simulated and real-world performance in robotics problems \citep{parmar2021fundamental}. 
Furthermore, velocity measurements are extremely sensitive to the measured time, as they change almost instantaneously during impacts. These issues become even more significant when learning a model of a real
system from noisy sensor measurements. 
The smoothing effect  of deep neural networks is therefore particularly harmful for modeling impacts and collisions. This is especially significant due to the sparsity of reliable data points that can be collected during and around impact which is pronounced by the lower reliability of an average sensor in these regions. \citet{parmar2021fundamental} provides an excellent discussion on the main challenges of using deep learning systems (as well as analytical methods) for modeling contact dynamics, an important one being the degradation of model performance with increasing stiffness.

Learning to model contacts directly is therefore a difficult problem using both analytical and learning-based methods. However, instead of modeling contact dynamics directly, the robot can decouple the problem by first approximating the \textit{positions of contact points}, and then using a `safer' controller in the predicted regions. Static contact properties such as \textit{contact positions} and \textit{direction of impact} can be estimated more reliably (if the task plan is known beforehand) using either tactile/force-torque sensors or coarse depth images/point cloud models of the objects. Acquiring training samples for such measurements are also usually easier than performing an
analytical analysis of the interaction. In \citep{ugur2015bottom}, a robot is shown to acquire the training samples' labels autonomously by interacting with objects to learn high-level rules for the objects that can be used for planning. Such interactive perception techniques have been used in several scenarios such as to estimate constraints or physical properties of objects \citep{katz2011factorization,barragan2014interactive}. The benefits of interactive perception are that they do not necessarily need pre-training and they help disambiguate between scenarios as well as in observing otherwise latent properties \citep{kroemer2019review}. For instance, a robot can figure out if an object is fixed or movable by pushing it. However, such methods usually require the robot to perform the task multiple times to build models or optimize model parameters, especially when trying to model dynamics properties of the interaction. In our contact-anticipation model (Section \ref{sec:probform-kf-pred}), we focus on interactively improving the knowledge of the robot about the locations of the contacts involved in the task using a Kalman-filter based update algorithm. These estimates of the locations of contacts are then used to define `transition' regions in the workspace of the robot, where it can expect contact-changes to occur, and therefore use a safer control strategy in such regions.

\subsection{Transition-phase controllers}

Using separate controllers is a common strategy for handling contact changes. Methods that use a transition-phase controller for changing-contact manipulation tasks focus on minimizing the discontinuities in the dynamics such that the controller is stable and tracks accurately after transition~\citep{mills1993control,marth1993stable,sidhik_learning}. However, these methods generally switch to a different controller only after a contact is detected, which can result in significant disruptions in the dynamics when the switch is made. These transitions results in the intake of large amount of energy into the system large spikes in forces and acceleration, and could potentially damage the robot or the domain objects. In contrast to these methods, our contact-change-handling module aims to predict the contacts and use it to transition to a safer `transition-phase' controller smoothly on-the-fly by interpolating between the control outputs \textit{before} collision occurs. 
\citet{hyde1997object} uses a transition-phase controller with constant low velocity to reduce impact effects in the guard regions of manipulation tasks modeled as hybrid systems. However, unlike their approach, our transition-phase controller modifies both velocity and stiffness to reduce impacts and vibrations during transitions, and can also automatically select the approach velocity to attain a desired impact force on collision (Section \ref{sec:velocity-profile}).


For the least delays and deviation from the original plan, the new approach velocity of the transition-phase controller should come into effect only when the robot is about to make a contact. Modifying the velocity requires modifying the timeline of the provided kinematic plan, hence causing deviation from the original plan (in terms of following the timeline) and sacrificing tracking accuracy. 
However, motion smoothness (motion derivatives at least up to jerk) can be guaranteed by making the transition trajectory continuous. To make the motion smooth in acceleration and jerk, a motion profile that is at least $C^4$ smooth is required. For kinematic time-optimal motion, different variants of trapezoidal velocity profiles are commonly seen \citep{biagiotti2008trajectory}. 
Several methods that create $C^4$ smooth trajectories using multiple trajectory segments have also been developed \citep{ahn2004arbitrary,nam2004study}. 
Several minimum-jerk motion profiles are also found in literature \citep{piazzi2000global,freeman2012minimum}. A trapezoidal $C^4$ smooth motion profile for point-to-point motion has been described in \citep{grassmann2018smooth} using a seventh-order $C^3$ polynomial function as the velocity profile. Such methods typically have many hyper-parameters that are tuned specifically for the task and requires knowledge on the limits of the system's jerk and higher order motion derivatives. In contrast, our framework uses a novel velocity profile for switching to the transition-phase controller, which is simpler in formulation with no additional hyperparameters and has continuous derivatives of all orders at every point of the function, making it $C^\infty$ smooth (Section \ref{sec:velocity-profile}). 


\section{Framework Description} 
\label{sec:framework}
Our overall framework takes as input the task-space target (i.e., desired) motion pattern and force control target profile for the robot to follow. In the absence of our framework, the robot would follow the provided task plan using a position controller, without compensating for the changing interaction dynamics as the task progresses. This means that the robot would have to use a high-stiffness controller within a contact mode so as to counter continuously varying environment dynamics. Similarly, there would be high discontinuities in the motion dynamics when the contact modes change (i.e. when the robot makes/breaks contact or when the environment changes discretely). High impacts during collisions could also damage the robot and/or objects involved. Our framework would ensure that these problems are avoided by modeling the piecewise continuous interaction dynamics during manipulation using a hierarchical learning framework, which uses an adaptive controller within a contact mode capable of modeling and compensating for continuously changing environments. The framework also learns to predict mode changes before they happen such that the robot can switch to a safer transition-phase controller during mode transitions such that the discontinuities during these transitions are reduced.


 \begin{figure}[tb]
  \begin{center}
    \includegraphics[trim={0cm 0cm 0 0},clip,width=\linewidth]{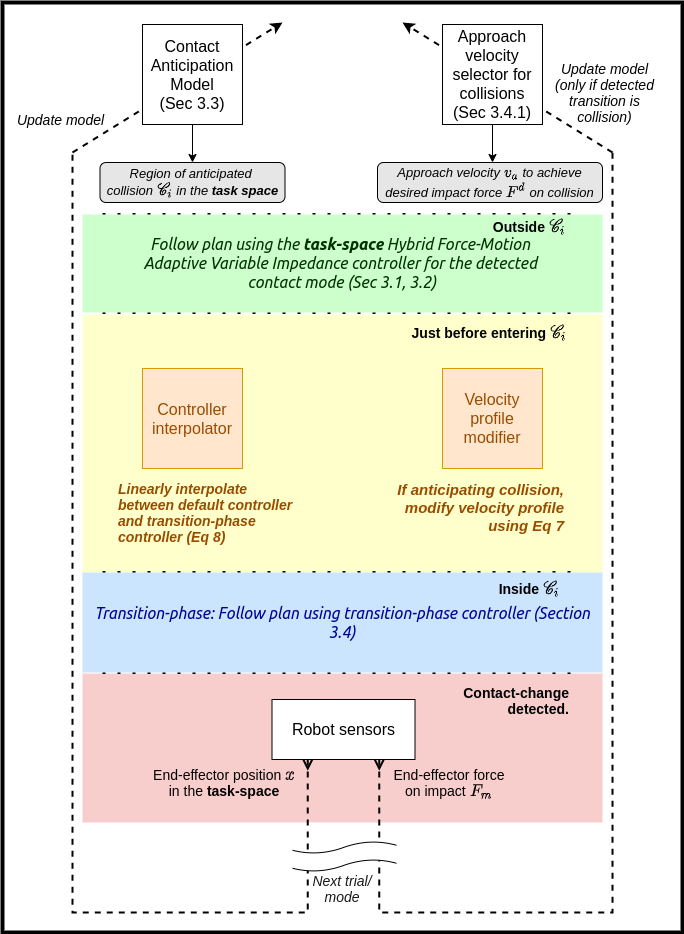}
  \end{center}
  \caption{Overview of the framework for smooth control of changing-contact manipulation tasks.}
  \label{fig:framework-diagram}
\end{figure}

Figure~\ref{fig:framework-diagram} presents an overview of our framework. The inputs are the desired motion trajectory, the force-torque sensor measurements, and the end-effector position. The default controller is the hybrid force-motion adaptive variable impedance controller (for hybrid dynamical systems) that is described in Sections~\ref{sec:probform-lowlevel-model} and \ref{sec:probform-moderecog}. The robot first tries to identify the contact mode that the robot is in and uses the appropriate forward model and controller for navigating that mode. The robot predicts the `region of anticipated mode-transitions` $\mathcal{C}$ in the task using the Kalman filter-based contact anticipation model described in Section~\ref{sec:probform-kf-pred}. The framework builds on this default controller's representation, enabling a task-space contact anticipation model that incrementally updates its contact prediction using a Kalman filter.  These predictions are used to minimize the time spent in the transition phase, and the controller parameters to be used in the transition phase is set adaptively to achieve a smooth motion profile (and a desired impact force if collisions are expected), as explained in Sections~\ref{sec:transition-ctrlr} and (Section~\ref{sec:smooth_transition}). Once the transition is completed using a suitable controller to minimize discontinuities, the robot identifies the new mode, uses the appropriate default controller and revises the parameter values suitably.

\subsection{Control Strategy and Learning Dynamics in a Mode}
\label{sec:probform-lowlevel-model}
For any given task, the desired motion trajectory $P$ is provided as a sequence of mappings from time to the end-effector pose and force (for force control). It is obtained through a single demonstration of the task by a human moving the manipulator. Since the controller operates in the Cartesian-space, the trajectory is in the form of segments that are (each) assumed to be smooth, continuous, and jerk-free. Transition between segments is accompanied by a change in the direction of (force, motion) control, and $P$ does not account for the collisions (i.e., contact points).



The basic controller in our framework follows the standard impedance control formulation with a force control term, making it a hybrid force-impedance controller:

\begin{equation}
    \bm{u}_t = \bm{H}_t  + \mathbf{K}_t^\mathbf{p}\Delta\bm{x}_t + \mathbf{K}_t^\mathbf{d}\Delta\bm{\dot{x}}_t + \bm{u}^\mathbf{fc}_t + \bm{u}^\mathbf{ff}_t \label{eq:basic_ctrl}
\end{equation}
where $ \bm{u}_t $ is the robot's control command (i.e., task space force) at time $t$, term $\bm{H}$ denotes the other dynamics compensation terms (inertia, Coriolis and gravity); $\mathbf{K^p}_t$ and $\mathbf{K^d}_t$ are the (positive definite) stiffness and damping matrices of the feedback controller for motion; $\bm{u}^\mathbf{fc}$ is force feedback control command to achieve orthogonal force targets if desired; and $\Delta\bm{{x}}$ and $\Delta\bm{\dot{x}}$ are the errors in the end-effector position and velocity at each instant. In the absence of external disturbances, the feed-forward term $\bm{u}^\mathbf{ff}_t$ can be zero. However, if there are external wrenches acting on the end-effector, a good forward model could be used to provide appropriate values for $\bm{u}^\mathbf{ff}_t$ that can help the robot follow $P$ accurately without being affected by those disturbances.

In our formulation, the forward model of each contact mode is learned as the robot attempts to follow $P$ for the given task. To avoid explicit dependence of the forward model on time, a Gaussian Mixture Model (GMM) is fit over points of the form $[\bm{S}_{t-1}, \bm{D}_t]$, where $\bm{S}_t$ can be any combination of features that uniquely represent the robot's state for the task, and $\bm{D}_t$ is interaction effects felt by the robot at its end-effector at time $t$.  $\bm{S}_t$ can contain information about end-effector pose ($\bm{x}_{t}$), velocity ($\dot{\bm{x}}_{t}$), forces ($\bm{F}^{ee}_{t}$), etc. while $\bm{D}_t$ would be measurable interaction effects such as end-effector forces ($\bm{F}^{ee}_{t}$) and torques ($\bm{\tau}^{ee}_{t}$).

In our framework, we aim to predict the end-effector forces and torques from previous measurements of end-effector velocities and wrenches. However, instead of their 3D vector representations we use the \textit{magnitudes} of force, torque, and end-effector
velocity (linear and angular separately) for learning and prediction. Since the magnitudes of frictional forces and torques are independent of the direction of motion (for objects having
consistent friction properties) and depend only on the relative speed
of motion of the objects, such simplified representation is sufficient
to learn and predict the end-effector forces and torques along the
direction of motion. This reduced representation of forces and torques
makes the learning process simpler, more computationally efficient, and
also \textit{independent of the direction of motion}. The learned
model always predicts the forces and torques along (or against) the
direction of motion.  Since the end-effector's direction of motion is
always known, the components of force and torques along the axes of
motion can be recovered when needed. The final state space where the forward model is learned can therefore be represented as $X_t = [\bm{S}_{t-1}, \bm{D}_t]$, with:

\begin{align}
    \bm{S}_{t-1} &= \left[\|\dot{\bm{x}}_{t-1}^{lin}\|, \|\dot{\bm{x}}_{t-1}^{rot}\|, \|\bm{F}^{ee}_{t-1}\|, \|\bm{\tau}^{ee}_{t-1}\|\right] \\
    \bm{D}_t &= \left[\|\bm{F}^{ee}_{t}\|, \|\bm{\tau}^{ee}_{t}\|\right]
\end{align}



The predictions from the forward model would then provide the feed-forward term $\bm{u}^\mathbf{ff}_t$ of the controller (Eq. \ref{eq:basic_ctrl}) that cancels out the effect of the predicted wrenches during motion, providing the control equation:
\begin{subequations}
\begin{align}
  &\bm{u}_t = \bm{H}_t  + \mathbf{K}_t^\mathbf{p}\Delta\bm{x}_t + \mathbf{K}_t^\mathbf{d}\Delta\bm{\dot{x}}_t + \lambda_{t-1} \bm{W}_t^{pred} + \bm{u}^\mathbf{fc}_t \label{eq:final_control_eq} \\
  &\mathbf{K}_{t}^\mathbf{p} = \mathbf{K}_{free}^\mathbf{p} + (1-\lambda_{t-1})(\mathbf{K}_{max}^\mathbf{p} - \mathbf{K}_{free}^\mathbf{p}) \label{eq:kp_update} \\
  &\lambda_t = 1 - \frac{1}{1+e^{-r(\varepsilon_t - \varepsilon_0)}}
  \label{eq:logistic_func}
\end{align} \label{eq:ctrl_eqns}
\end{subequations}
where $\lambda_{t-1} \bm{W}_t^{pred}$ is the weighted feed-forward wrench (end-effector forces and torques) predicted by the forward model associated with the present mode $m_t$. The factor $\lambda_t$, a function of the accuracy of the forward model at instant $t$, maps the error in prediction from the forward model ($\varepsilon_t$) to a value between 0 and 1, e.g., logistic
function in Equation \ref{eq:logistic_func}. The logistic growth rate $r$ and the Sigmoid midpoint $\varepsilon_0$ are hyper-parameters tuned for the task. This control law formulation relies on the feed-forward term only if the forward model's predictions are found to be accurate; if not, the feedback control term is prioritized.
Equation~\ref{eq:kp_update} describes the adaptation of stiffness parameters as a function of the prediction accuracy. $\mathbf{K}_{max}^\mathbf{p}$ is the maximum allowed stiffness, and
$\mathbf{K}_{free}^\mathbf{p}$ is the minimum stiffness for accurate position tracking in the absence of external disturbances (i.e., free space motion). The damping term is updated as
$\mathbf{K}^\mathbf{d}_t = \sqrt{\mathbf{K}^\mathbf{p}_t/4}$ considering the manipulator as a critically-damped systems. This way, the robot can follow $P$ accurately using high feedback gains if the forward model is unreliable, but can be compliant if they are compensated well by the feed-forward term.

To incrementally update the forward model of our framework using new measurements during task execution, we used an online variant of GMM called the Incremental
GMM (IGMM)~\citep{Song2005,engel2010incremental,ahmadincremental}. IGMM can update model parameters and incorporate additional components to the mixture model using a measure of closeness and frequency which are defined by the user using hyperparameters. For more information regarding the incremental algorithm used, readers are referred to \citep{engel2010incremental}. IGMM internally uses a variant of the Expectation-Maximization (EM)
algorithm to fit the model and maximize the following likelihood
function:
\begin{align}
  L(\theta) = p(\mathbf{X}|\theta) = \prod_{n=1}^T p(X_n | \theta) =
  \prod_{n=1}^T \left[ \sum_{j=1}^M p(X_n | j)p(j) \right]
\end{align}
where $\theta = (\mu_j, \sigma_j, p_j) \text{ for } j = 1...M$ are the
parameters of the $M$ components of the GMM. $\mathbf{X} = (X_1, ...,
X_T)$ represents the points to be fit, with $X_t = [\bm{S}_{t-1}, \bm{D}_t]$.
Each point contains information about the \textit{previous}
end-effector state, along with the \textit{current} wrench.
Once trained, the forward model provides a function:
\begin{align}
    f_\text{fm}: \bm{S}_t \mapsto \bm{D}_{t+1} \label{eq:w1-fmodel_function}
\end{align}
that predicts $D_{t+1}$ at the next time step as a function of the current (measured) value of $\bm{S}_t$, using Gaussian Mixture Regression (GMR)~\citep{sung2004gaussian}. The incremental nature of our dynamics model allows the robot to capture the smoothly varying interaction dynamics between the robot end-effector and the environment within a contact mode.
\subsection{Contact Mode Recognition and Identification}
\label{sec:probform-moderecog}
In a typical manipulation task, the interaction dynamics undergoes a non-smooth transition when a contact change occurs, moving the state of the hybrid piecewise continuous system to a new contact mode. Our approach for recognizing known modes and identifying new ones is based on the observation that any change in mode is accompanied by a sudden significant change in the sensor readings. In our framework, the robot responds to pronounced changes in force-torque measurements by briefly using a high-stiffness control strategy while quickly obtaining a batch of sensor data to confirm and respond to the transition. The robot learns a new dynamics model if a new mode is detected, and transitions to (and revises) an existing dynamics model if the transition is to a known mode.

The key factor influencing the reliability and generalizability is the choice of feature representation for the modes. This representation is task-dependent and should be able to uniquely identify the different contact modes in the task. 
For the task of sliding an object over surfaces with different values
of friction (Figure \ref{fig:example}), for instance, the property that strongly influences the end-effector
forces ($F^{ee}$) is the friction coefficient between the object and
the surface. When two objects slide over each other at constant
velocity, $F^{ee}$ is proportional to the applied normal force ($R$)
and the friction coefficient ($\mu$) (assuming the relative
orientation of their surface normals do not change); $\mu$ can then be
estimated as:
\begin{equation}
  \mu \propto \frac{\lVert F^{ee}\rVert}{{R}} \label{eq:mu_f_R}
\end{equation}
A concise feature representation for this task is thus $\frac{\lVert
  F^{ee}_t\rVert}{{R_t}}$, which has the effect of making mode
classification independent of the magnitude of the applied force.


The management of modes is based on an online incremental clustering algorithm called Balanced Iterative Reducing and Clustering using Hierarchies (BIRCH)~\citep{zhang1997birch} in the Scikit-learn library~\citep{scikit}. This algorithm incrementally and clusters incoming data for given memory and time constraints, without examining all data points or clusters. Each cluster is considered to represent a mode in an abstract feature space, with the clusters updated using batches of the feature data. The fraction of the input feature vectors assigned to any cluster determines the confidence in the corresponding mode being the current mode. If the highest such confidence value is above a threshold, the dynamics model of that mode is used and revised until a mode change occurs. If the feature vectors are not sufficiently similar to an existing cluster, a new cluster (i.e., mode) and the corresponding
dynamics model are constructed and revised (Section~\ref{sec:probform-lowlevel-model}) until a mode transition occurs. 

\begin{algorithm}[htb]
  \algsetup{linenosize=\normalsize}
  \normalsize

  \caption{\textbf{Control loop for hierarchical model learning of piecewise-continuous dynamics}}
  \label{alg:control}
  
  \DontPrintSemicolon 
  
  \SetKwInOut{Input}{Input}
  
  \SetKwInOut{Output}{Output} 

  \Input{Desired motion pattern as sequence of task space way-points, Control parameters: $\mathbf{K}^p_{free}, \mathbf{K}^p_{max}$; Dynamics models corresponding to modes $\mathcal{M}= \{{f}_i: i\in [1, N]\}$; Current mode: $m=0$.}  

  \BlankLine

  \While{Motion pattern not complete} {


      \If{mode transition detected}{

        \tcp*[h]{Set high stiffness}
        
        $\mathbf{K}^p_{t} \leftarrow \mathbf{K}^p_{max}$\;
        
        \tcp*[h]{Detect (new/existing) mode}
        
        $m$ = detect\_classify\_mode() %
        
        \tcp*[h]{Populate new model for new mode}
    
        \If{new mode found}{
            $\mathcal{M} = \mathcal{M}\cup f_m$
        }
      }

      
        Update $f_{\text{fm}|m}$ online and use it for control in the identified mode (Eq. \ref{eq:ctrl_eqns})
    }
    
\end{algorithm}

Algorithm~\ref{alg:control} is an overview of the framework's control loop for a changing-contact manipulation task, e.g., sliding an object on a surface. It proceeds until a desired motion pattern is completed. The mode-detection module is triggered when there is a contact change is detected by the force-torque sensor (line 2). The robot detects mode changes when there are substantial changes in the sensor measurements. The robot responds by setting a high stiffness (line 3), collecting sensor measurements, determining the transition to a new or existing mode (line 4), and creating new models if necessary (lines 5-7). In the absence of a mode transition (e.g., the detected change in sensor measurement was an anomaly), the robot continues with the current mode and dynamics model (line 9). 
\subsection{Predicting contact changes}
\label{sec:probform-kf-pred}

In a static environment, a manipulator robot may experience discontinuities in dynamics (contact changes) of two types: (i) impact transitions (collisions), and (ii) impact-less transitions. 

Collisions occur when a moving robot (end-effector) comes in contact with a fixed object in its workspace. Given that a task plan is known in a fixed static environment, there are several methods of estimating the contact locations for such mode changes. Unless occluded or visually indistinguishable, visual sensors are prime candidates for providing rough initial estimates of where collisions can occur in a planned motion. An important characteristic of collision-based mode change is that it will always remove at least one degree of freedom (DoF) of the robot in a functional coordinate space. Collisions are marked by large spikes in force-torque measurements as a function of the relative velocity of approach as well as material properties such as coefficient of restitution and hardness of the objects involved. Furthermore, collisions also bring about a discontinuity in the velocity of the end-effector by forcing at least one dimension in the functional coordinates to go to zero. This causes spikes in acceleration, jerk etc. All these effects can be attributed to the sudden loss in a degree of freedom in the functional coordinate space. 

The second type of contact change occurs when the interaction discontinuities occur due to discrete changes in the environment dynamics that are not due to impact. A simple example is when robot slides across two surfaces whose friction are different (Figure \ref{fig:exp-impactless_task}); here, the robot experiences sudden changes in frictional resistance as it crosses the boundaries between the surfaces, but the transition does not necessarily cause large impact forces or drop in velocity. The main feature that distinguishes impact-less transitions from collisions is that these mode transitions occur without a loss in degree of motion freedom.

Anticipating contacts (collisions) by predicting impact forces or time to collision is challenging because these parameters are influenced by robot dynamics and controller parameters, e.g., reducing the velocity or stiffness reduces the impact force and increases time to contact. 
\textit{Static} contact parameters such as end-effector position during impact and direction of contact force can be predicted more reliably; they do not change significantly for task repetitions if we can make the reasonable assumption that the trajectory and environmental attributes do not change significantly between repetitions.

In our contact anticipation model, the robot's belief about the position of each expected contact location in the assigned motion trajectory is modeled as a multivariate Gaussian, with the covariance ellipsoid denoting the uncertainty and the ``region of anticipated mode transition" $\mathcal{C}$. In the context of this work, a `contact position/location' refer to a position in the task-space of the robot where a mode switch occurs when following the provided task plan. Therefore, in any trial of the task, the robot expects a specific contact position $\mathbf{c}$ to lie within the corresponding $\mathcal{C}$.
Since the controller is in the task space, each contact location's representation is compact and is updated over very few trials of the task using a Kalman filter with the state update equation: $\mathbf{\dot{c}} = \mathbf{A}\mathbf{c} + \mathbf{B}\textbf{u}_k + w$, where $\mathbf{c}$ is the contact position, $\mathbf{A}$ is the object's self-activation ($I$ for positively activated objects), $\mathbf{B}$ is the control matrix capturing the effect of action $\textbf{u}$ on contact position, and $w$ is Gaussian noise modeling the uncertainty in the contact location.
The sensor model uses the end-effector pose (obtained by forward kinematics with joint positions) as measurement when a contact is detected; noise in the sensor model depends on the joint encoder noise and forward kinematics. The corrected estimate of the contact point results in a reduced covariance ellipsoid for subsequent trials.
Although this representation supports contact with movable objects, we assume in this paper that the end-effector only makes contact with stationary objects.
These simplifications result in Gaussian updates using the noisy measurements based on the robot's kinematics model (since sensor input is from an FT sensor) each time the robot experiences a contact change.

\subsection{Transition-phase controller for contact changes} 
\label{sec:transition-ctrlr}
Next, we describe the transition controller for safe and smooth dynamics during mode transitions. The transition controller follows the same control structure as our default variable impedance controller (Eq. \ref{eq:final_control_eq}) and varies only in the choice of the control parameters. Depending on the type of mode transition (collisions or impact-less), the choice of the control parameter values can vary, but the basic structure remains the same:

\begin{equation}
    \bm{u} = \mathbf{K}^\mathbf{p*}\Delta\bm{x} + \mathbf{K}^\mathbf{d*}\Delta\bm{\dot{x}} + \bm{u}^\mathbf{ff} + \bm{H} \label{eq:transition_ctrl_law}
\end{equation}
where $\mathbf{K}^\mathbf{p*}$ and $\mathbf{K}^\mathbf{d*}$ are the stiffness and damping parameters of the transition-phase controller that can vary depending on the type of mode transition. The different desired properties of the transition-phase controller for the two types of mode transitions are described in the next sections.

\subsubsection{Transition-phase controller for collisions:}

Since the permitted impact force may differ based on the task, e.g., large forces can damage delicate objects, we imposed a limit on the maximum allowed impact force.
Also, we figured out experimentally that reducing the controller stiffness helps reduce the jerk in motion after impact by providing compliance, but has no significant effect on impact forces because the error and stiffness term in the feedback control loop come into effect only after contact is made. A safe controller should thus have lower stiffness for reducing vibrations. In addition, the approach velocity was observed to be directly proportional to the impact force, especially when the robot registered a contact while moving in free space. We exploited these insights in our framework and used a simple linear (regression) model to capture the relationship between impact force and approach velocity between a pair of objects. This model was then used to compute the approach velocity for any (given) desired force on impact. 



Since the robot may not initially have a model of the (linear) relationship discussed above, it starts with a safe low velocity during the first trial of any given task and a target force on impact. It then uses the difference between the target and measured force on impact to revise the approach velocity for the next iteration of the task: 
\begin{align}
\Delta v_{a} = \beta (F_{d} - F_{m}) \label{eq:vel_update}
\end{align}
where $\Delta v_a$ is the change in approach velocity, $F_d$ is the desired impact force along motion direction, $F_m$ is the measured impact force, and $\beta$ is a learning rate that is ideally less than or equal to the slope of the function relating impact force to approach velocity. Over time, this method enables the robot to learn a task-specific approach velocity for a desired impact force. 

\subsubsection{Transition-phase controller for impact-less mode transitions:}

If the robot expects an impact-less transition in $\mathcal{C}$, it does not have to reduce its velocity and can continue following the original plan. Reducing velocity would bring delays in the task which is undesirable. However, having a high-stiffness controller would help in identifying new modes after transition. This is because high stiffness ensures that the robot motion is steady and the sensor measurements are more accurate, which helps in mode identification. For this reason, the ideal transition-phase controller for impact-less transitions should switch to a high-stiffness controller. The parameter values for this transition phase controller (Eq. \ref{eq:transition_ctrl_law}) in our framework is set to use the maximum allowed values of the active adaptive variable impedance controller (base controller), i.e., $\mathbf{K}^\mathbf{p*}=\mathbf{K}_{max}^\mathbf{p}$.
\subsection{Smooth Transition between Controllers} \label{sec:smooth_transition}
The robot will have to switch to a transition-phase controller smoothly as it enters the region of anticipated mode transition $\mathcal{C}$.
To avoid discontinuities in motion dynamics, the robot needs to smoothly transition from a normal (base) controller with output $\mathbf{u_1}$ to the appropriate transition-phase controller with output $\mathbf{u_2}$. Linear interpolation of $\mathbf{u_1}$ and $\mathbf{u_2}$ over a time window $[0,T]$ is possible since they are of the same task-space representation, and can produce smooth transition between these controllers:
\begin{align}
  \mathbf{u} = &(1 - \alpha)\mathbf{u_1} + \alpha \mathbf{u_2};\qquad
  \alpha = t/T \qquad t \in [0,T]\label{eq:ctrl_interp}
\end{align}
where $T$ is the desired duration of the transition between the controllers. As long as the outputs from the two controllers ($\mathbf{u_1}$ and $\mathbf{u_2}$) are individually smooth, the output of the combination will also be smooth. In this work, controllers use the task-space representation described earlier, with $\mathbf{u_2}$ being the output of the fixed transition-phase controller (low- or high-gain depending on the transition type) as the arm approaches the contact point. We use this strategy to smoothly transition between controllers of different stiffnesses such that the transition is completed by the time the robot reaches $p_c$. A similar approach is used to smoothly transition from the transition-phase controller to a normal controller after contact is made.


\subsubsection{Velocity Profile Shaping:}
\label{sec:velocity-profile}
Recall that transition-phase controllers for handling collisions use a lower velocity than that used in the original kinematic sequence $P$ to reduce the force on impact. Also, the switch to this controller will take place at different points in the trajectory as the region $\mathcal{C}$ is revised over time. The trajectory's timeline thus has to be modified to account for the modified velocity profile. To achieve this objective, we enable the robot to create a new velocity profile and time-mapping.
Our formulation results in motion that is smooth and continuous at all orders, i.e., is $C^\infty$ smooth.

\begin{figure}[tb]
  \begin{center}
    \includegraphics[trim={0cm 0cm 0 0.7cm},clip,width=\linewidth]{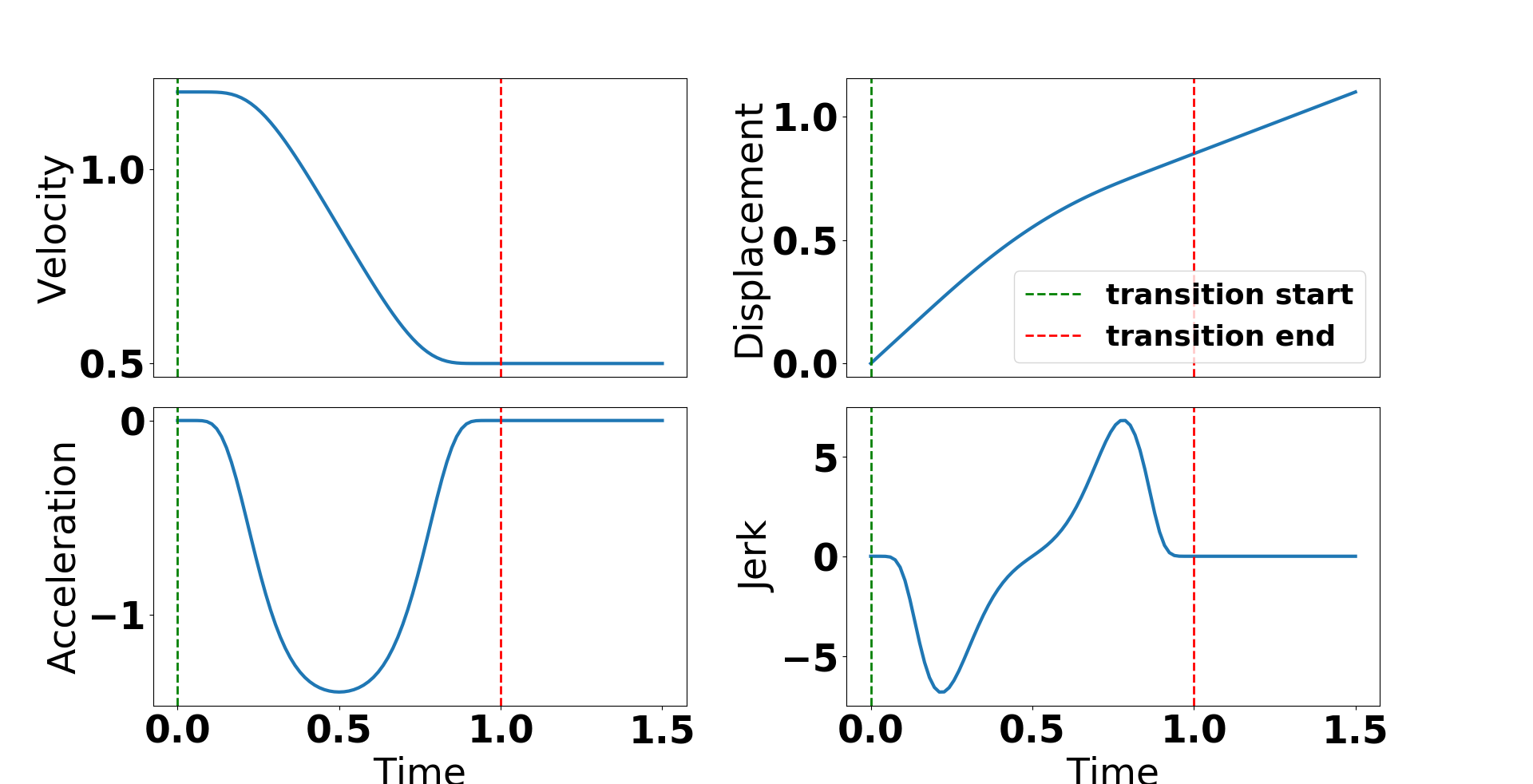}
  \end{center}
  \caption{Velocity plots with matched position, acceleration, and jerk plots. Velocity varies from 1.2 to 0.5 in unit time.}
  \label{fig:velprofeg}
\end{figure}

Without loss of generality, assume that $P$ is along one dimension with velocity $v_1$. Assuming that transition starts at time $t_1$ with $v_1$ and has to be completed at $t_2$ with velocity $v_2$ as the robot crosses boundary point $p_c$ of $\mathcal{C}$, our velocity profile is defined as:
\begin{equation} v(\tau) = \begin{cases}
    v_1 + \frac{(v_2 - v_1) e^{-1/\tau}}{e^{-1/\tau} + e^{-1/(1-\tau)}} &\quad\text{if }0 < \tau < 1,\\
    v_1 & \quad\text{if }\tau \leq 0\\
    v_2 & \quad\text{if }\tau \ge 1\\
  \end{cases} \label{eq:vel_prof}
\end{equation}
where $\tau = t/T = t/(t_2 - t_1)$. For $\tau \in (0, 1)$, $e^{-1/\tau}$ has continuous derivatives at all orders at every point $\tau$ on the real line. Since $v(\tau)$ has a strictly positive denominator for all points in its domain and velocity limits are enforced $\forall \tau \notin [0, 1]$, this profile provides a smooth transition from $v_1$ to $v_2$ over $[t_1, t_2]$ and $v(\tau)$ is continuous despite its piece-wise definition. Acceleration and jerk are computed as first- and second-order derivatives of $v(\tau)$ with respect to $\tau$, and position trajectory is obtained by integrating the profile; all motion derivatives are continuous -- see Figure~\ref{fig:velprofeg}.

\section{Experimental evaluation} 
\label{sec:expres}
In this section, we experimentally demonstrate our framework's capabilities. We first evaluate the effectiveness of the overall framework in producing smooth overall motion in the presence of collisions and impact-less mode transitions in Section \ref{sec:exp-smooth_motion}. Then we proceed to evaluating the different components of the framework in performing their separate functions by comparing with baselines from literature wherever possible. For this, we demonstrate first the need for having an incrementally updating model for control in continuously and smoothly varying environments (Section \ref{sec:need_for_inc_models}). We then show the advantage of our adaptive variable impedance controller in such environments (continuous contact) by comparing it with a few other adaptive control strategies (Section \ref{sec:ctrl-compare}). We then demonstrate experimentally the ability of the hierarchical model learning framework to identify and model different contact modes in tasks where the interaction dynamics changes discretely (Section \ref{sec:exp_hyb_framework}). Section \ref{sec:exp-hyb-sys} experimentally justifies the choices made in the formulation our hybrid framework for handling piece-wise continuous dynamics, by comparing performance with a sophisticated, long-term prediction algorithm as the baseline in a 3D simulation domain involving a 7-DoF robot arm.

Our framework is tested in the context of a physical 7 DoF robot performing a manipulation tasks that involves multiple contact changes (collisions and/or impact-less mode transitions). Comparisons with baseline methods are set up in simulation domains where the interaction dynamics can be controlled and repeated. 
In most experiments, the performance measure is the accuracy or error in achieving the desired trajectory along with qualitative measures such as smoothness of interaction dynamics and delays in task completion.

The primary hypotheses being tested in this section are:

\begin{itemize}

\item[\textbf{H1:}] Incrementally updating adaptive models are required to capture smoothly changing and/or unmodelled interaction dynamics in a contact mode. This helps our adaptive variable impedance controller perform better than traditional adaptive control strategies in such environments;

\item[\textbf{H2:}] Our hybrid hierarchical model learning strategy for piecewise-continuous dynamics detects and models contact modes in a reliable and efficient fashion;

\item[\textbf{H3:}] Corrective system feedback in the task-space and online adaptation of hybrid models (as in our hierarchical model learning) is required for effectively capturing piecewise-continuous interaction dynamics in changing-contact manipulation.

\item[\textbf{H4:}] The contact-change-handling module of our framework reliably predicts contact changes and provides smooth motion during mode transitions;
\end{itemize}

\subsection{Incremental models for continuously changing dynamics} 
\label{sec:need_for_inc_models}
Our prior work on smooth control for continuous dynamics has been evaluated on a seven degree of freedom (DoF) robot arm performing  continuous contact tasks~\citep{humanoids_varimp}.
To demonstrate the importance of incrementally adapting the forward model for continuous changes in the environment (\textbf{H1}), we compare our incrementally-trained forward model with a fixed GMM model pre-trained using the same data. We used a custom-built 2D simulated environment (Figure \ref{fig:exp-ctrl-compare-aic-lag-screenshot}) based on Box2D physics engine~\citep{catto_2017} with noiseless measurements.

\begin{figure}[t]
\centering
\begin{subfigure}{\linewidth}
  \centering
  \includegraphics[width=\linewidth]{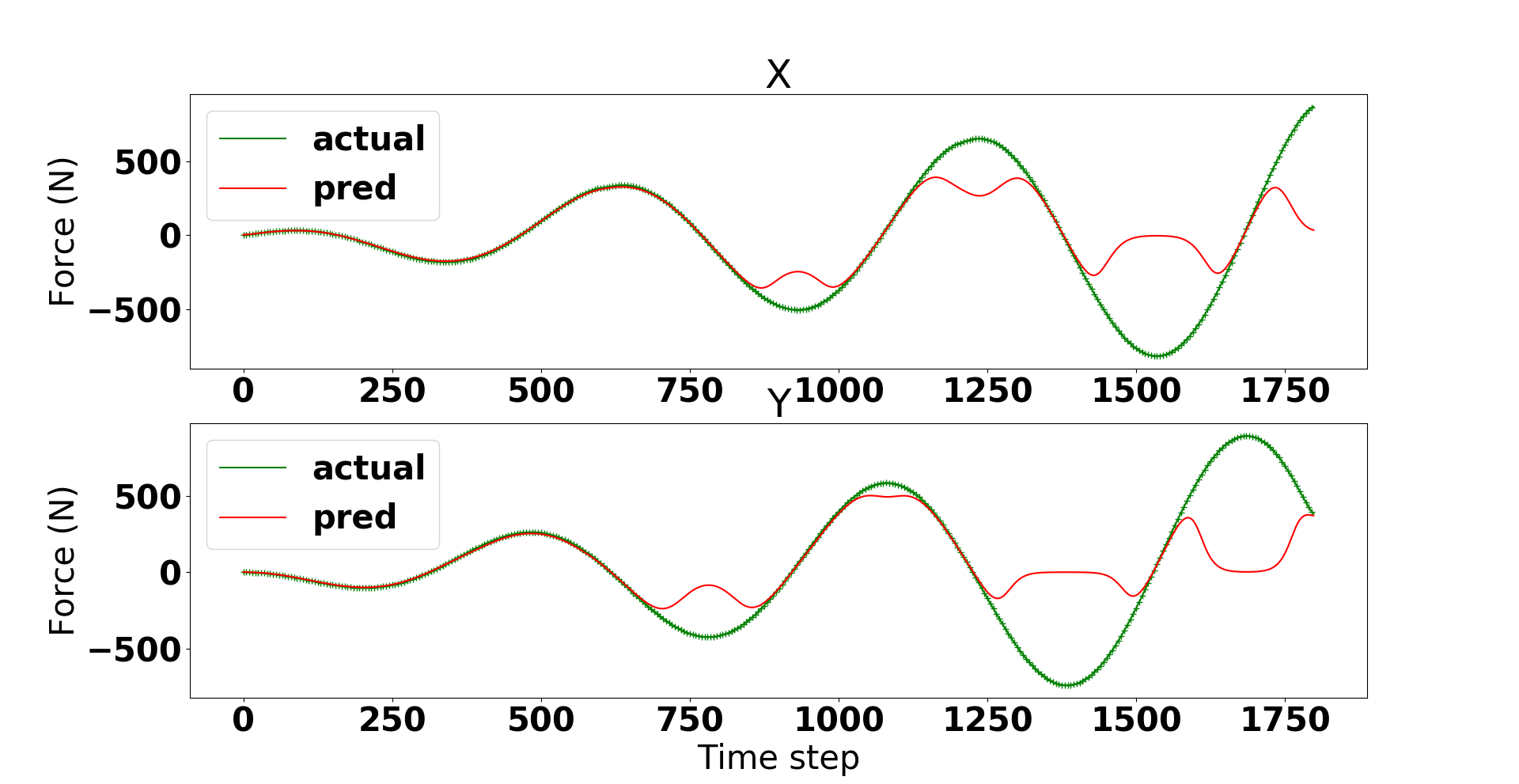}
  \caption{Predictions from fixed GMM forward model.}
  \label{fig:exp-inc-compare-porridge-no-inc}
\end{subfigure}%

\begin{subfigure}{\linewidth}
  \centering
  \includegraphics[width=\linewidth]{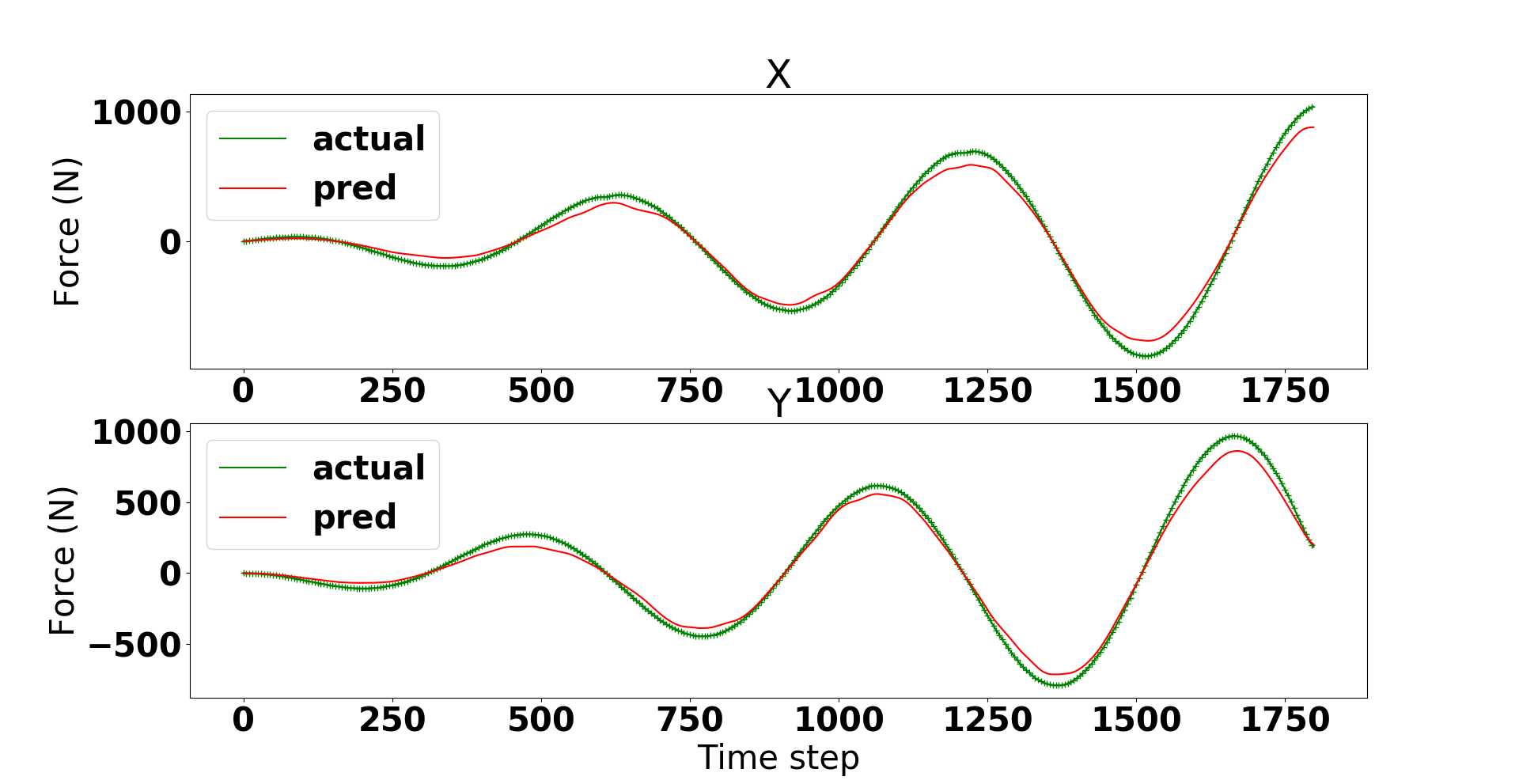}
  \caption{Predictions from incremental GMM model.}
  \label{fig:exp-inc-compare-porridge-inc}
\end{subfigure}
\caption{Comparison of fixed and incremental GMM models in ``porridge'' environment; \textit{red:} predicted values; \textit{green:} actual measurements.}
\label{fig:exp-inc-compare-porridge}
\end{figure}

We simulated a porridge-stirring task in simulation with the viscosity continuously increasing as the robot moves. 
The robot was made to move along a circular path in an environment where viscosity increases continuously from $0~Ns/m^2$ to $80~Ns/m^2$ in steps of $0.1~Ns/m^2$ every simulation step (the total circular trajectory had 600 steps). The measurements from this experiment were used to build a fixed GMM model for one-step prediction of end-effector forces from previous end-effector velocity and force. 
This model's predictions in an environment where the viscosity increases from $0-150~Ns/m^2$ is shown in Figure \ref{fig:exp-inc-compare-porridge-no-inc}. We observe that the model is able to predict the forces accurately when the measurements are similar to the training data, but fails when the measured values are very different from the training data. Even when the predictions are accurate, it is largely due to the use of a constant high-stiffness controller. Although we do not show it here, when this learned model is used to guide the variable impedance behavior (in Eq. \ref{eq:kp_update}), it causes unreliable trajectory tracking. Also, for the pre-trained model to be useful, the training data will need to include samples from different regions of the state space; this imposes additional training time and memory requirements. These results indicate the need for using an incrementally learned forward model to account for continuously-changing interaction dynamics, and proves hypothesis \textbf{H1}. 



\subsection{Comparison with adaptive control methods} 
\label{sec:ctrl-compare}
To further test \textbf{H1}, we compared the capabilities of the adaptive variable impedance controller (base controller) of our framework with three adaptive control methods in the presence of continuously and smoothly varying interaction dynamics.


\subsubsection{Comparison with MRAC:} \label{sec:ctrl-compare-mrac}
We used an established MRAC method implementation as the baseline~\citep{131947}. 
Recall that MRAC methods attempt to adapt their control law based on a pre-defined reference model. The reference model in the chosen MRAC method uses a second-order spring-damper system for each dimension $i$, parameterized with user-defined values for desired natural frequency $\omega_i$ and damping ratio $\zeta_i$. The control law continuously adapts its control parameters to minimize tracking errors.

\begin{figure}[tb]
  \begin{center} 
    \includegraphics[trim={3cm 0cm 4cm 0},clip,width=\linewidth]{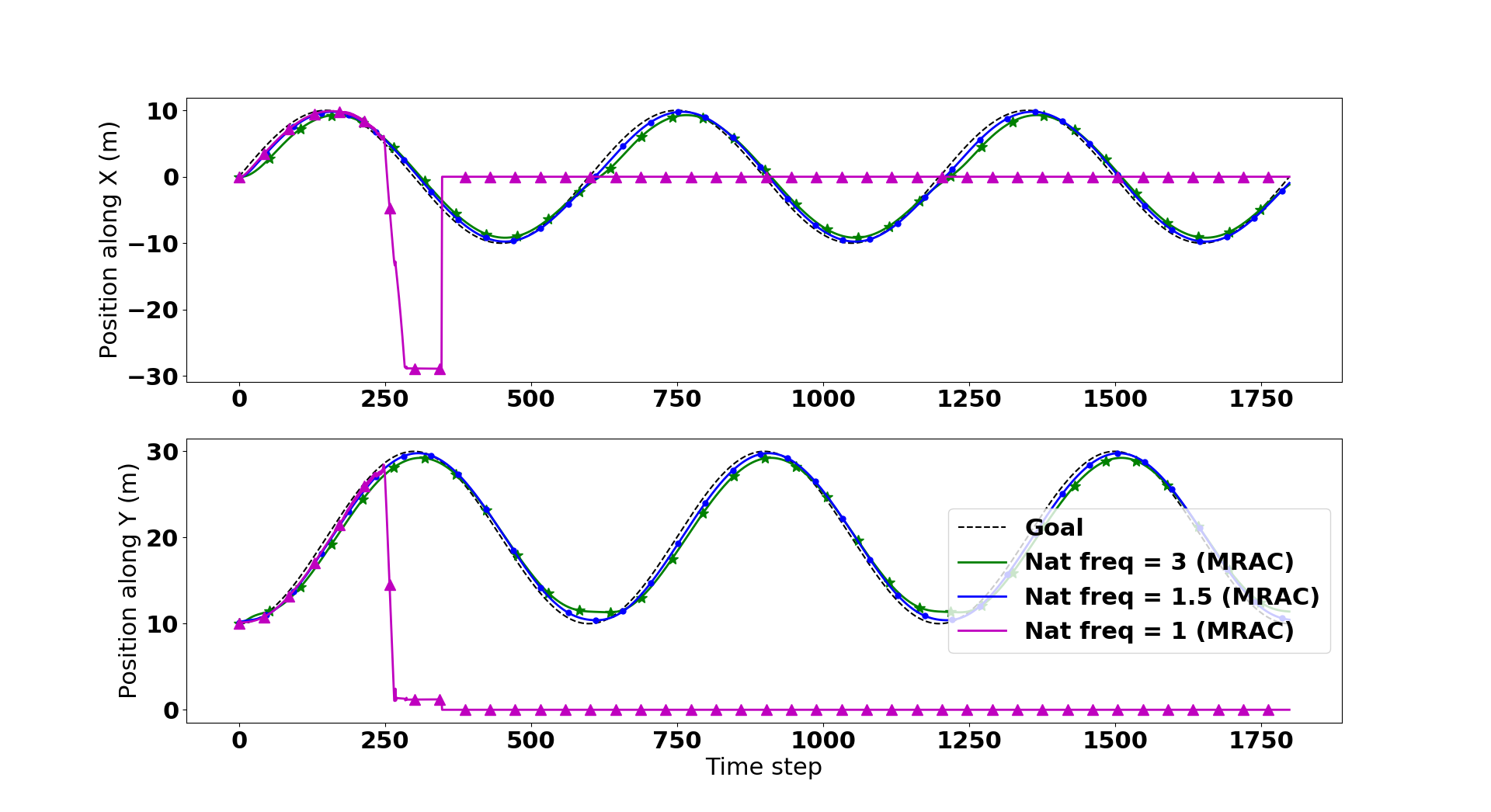}
  \caption{Performance of MRAC is highly dependent on the design of the reference model.}
\label{fig:exp-ctrl-compare-mrac-natfreq}
  \end{center}
\end{figure}

\begin{figure}[tb]
  \begin{center} 
    \includegraphics[trim={3cm 0cm 4cm 0},clip,width=\linewidth]{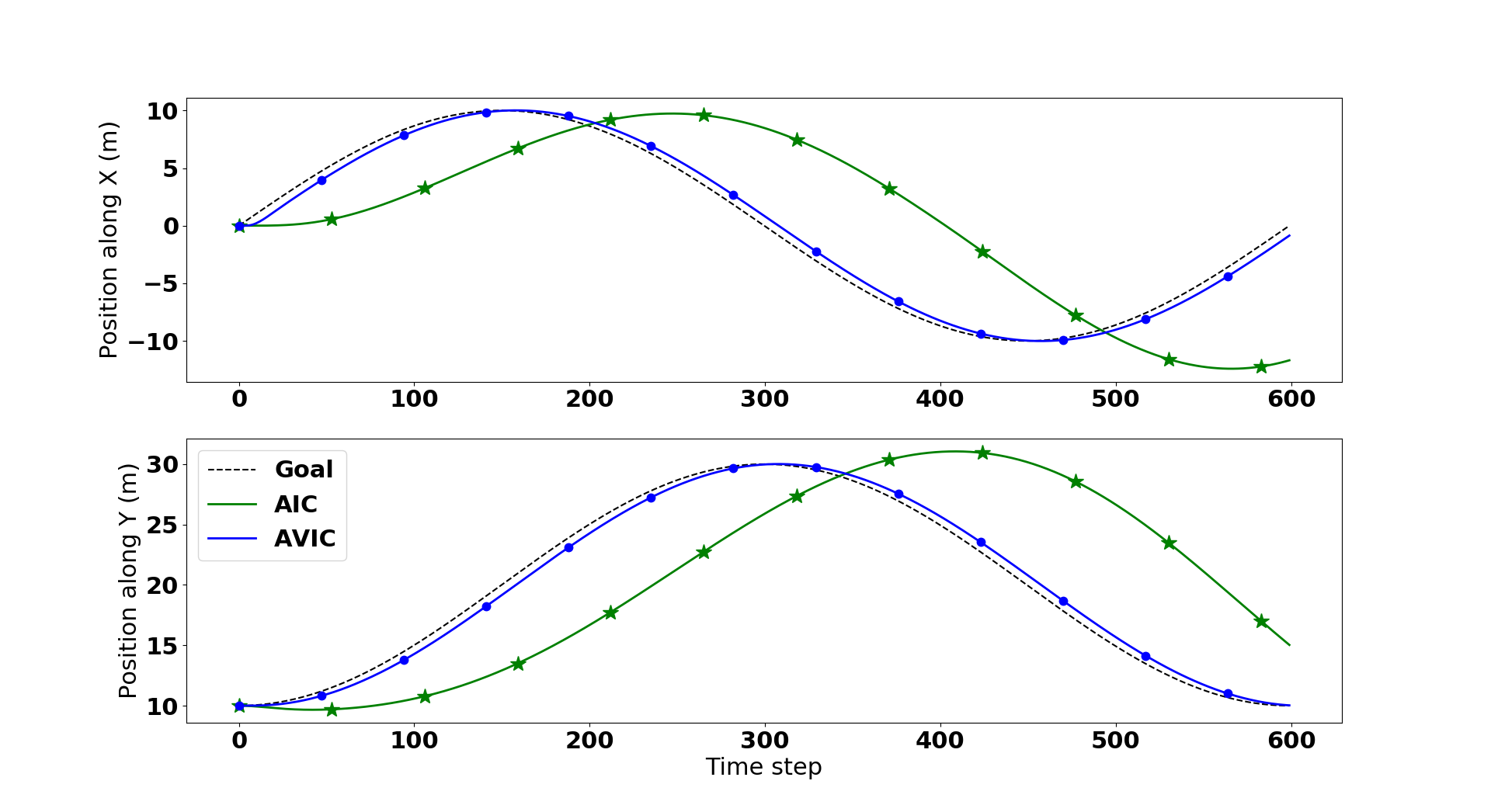}
  \captionof{figure}{Trajectory tracking comparison between AIC and our base controller.}
  
  \label{fig:exp-ctrl-compare-aic-lag-traj}
  \end{center}
\end{figure}


MRAC was tested in a ``multi-spring'' environment where the robot had to move along a circular trajectory while the end-effector is being pulled by three springs of different stiffness (Figure \ref{fig:exp-ctrl-compare-aic-lag-screenshot}). Tuning the large number of hyperparameters was found to be quite cumbersome for continuously varying environments, e.g., in Figure \ref{fig:exp-ctrl-compare-mrac-natfreq}, trajectory tracking is affected by the choice of the frequency parameter $\omega$ of the reference model, with stability affected adversely when value of $\omega$ is changed from 1.5 to 1. However, hyper-parameter selection for the ``porridge'' environment proved difficult and performance comparable to our framework could not be achieved using MRAC.



\subsubsection{Comparison with a self-tuning regulator:} 
\label{sec:ctrl-compare-aic}
As an example of a self-tuning controller, we chose the Active Inference Control (AIC) method~\citep{pezzato2020novel}. 
This adaptive control strategy supports continuous updates for changing environmental dynamics; the control parameters are iteratively updated to account for unaccounted energy in order to reach a fixed target. Given time, AIC converges to a fixed target smoothly and without any appearance of instability  for any dynamically changing environment. However, this is not useful when the robot has to follow a motion trajectory as the controller's parameters will typically not converge before the target changes to the next point in the trajectory.
As a result, the robot always lags behind the target with AIC (see Figure \ref{fig:exp-ctrl-compare-aic-lag-traj}), whereas our adaptive variable impedance controller compensates for external disturbances by directly predicting and canceling the forces, thus minimizing the lag in trajectory tracking.


\subsubsection{Comparison with a gain-scheduling controller:}

We implemented Biomimetic Adaptive Control (BAC), an RL-based approach that iteratively updates control parameters (gains) at each point in a repeating trajectory~\citep{yang2011human}. Given data from multiple trials of such a trajectory, parameters at each time step are updated using values from previous iterations at the same time step to minimize a cost function based on tracking error.


\begin{figure}[tb]
\centering

\begin{subfigure}{0.49\linewidth}
  \centering
  \includegraphics[width=\linewidth]{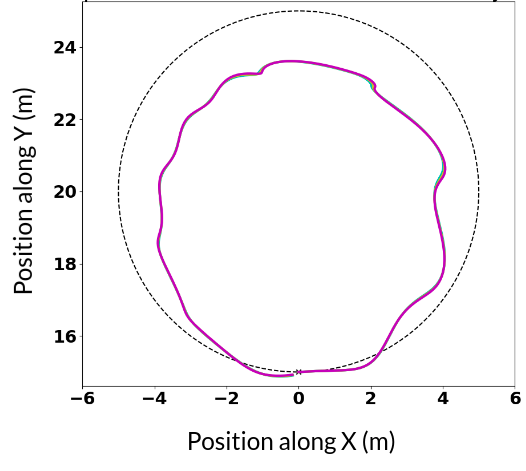}
  \caption{BAC is unstable with incorrect initialization of controller parameters.}
  \label{fig:exp-ctrl-compare-bac-instability}
\end{subfigure}%
\hfill
\begin{subfigure}{0.49\linewidth}
  \centering
  \includegraphics[width=\linewidth]{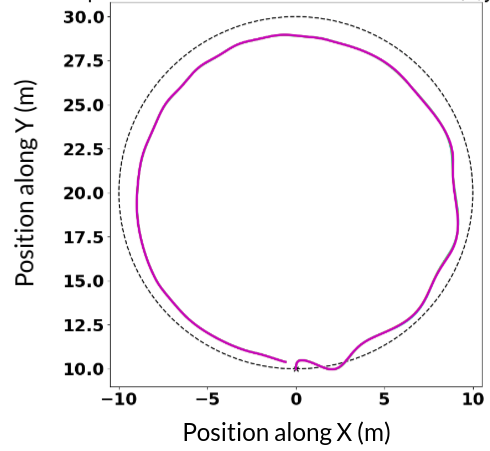}
  \caption{BAC parameters do not converge in 30 trials in ``multi-spring'' environment.}
  \label{fig:exp-ctrl-compare-bac-mspring}
\end{subfigure}
\vspace{-0.5em}
\caption{Trajectory followed using BAC by robot in trial 30 in two different experiments. Dotted line indicates target trajectory and solid line is the actual path followed by robot.}
\label{fig:exp-ctrl-compare-bac}
\end{figure}

In our experimental trials, we observed that bad parameter initialization causes instabilities and tracking irregularities that accumulate as the trajectory progresses, since BAC does not account for the temporal relationship between controller parameters within a trial. One example of such a trial with incorrect parameter initialization is shown in Figure \ref{fig:exp-ctrl-compare-bac-instability}. Due to this sensitivity to initialization, it is difficult to ensure smooth trajectory tracking when learning a time-indexed parameter profile.

Although BAC can provide good trajectory tracking for simple environments when good parameter initialization and enough trials are provided~\citep{yang2011human}, the parameters often do not converge in a limited number of trials with more complex environments such as the ``multi-spring'' environment in Figure \ref{fig:exp-ctrl-compare-bac-mspring}. Such methods also rely on the duration of the task (i.e., number of time steps in each trial) to be fixed across trials.

The results from these comparison experiments show that the incremental nature of the feed-forward model and the online adaptation of the feedback gains help our controller perform better than traditional adaptive control strategies in the presence of continuous and smoothly changing environments, supporting our hypothesis \textbf{H1}.


\subsection{Evaluating hybrid framework for piecewise continuous systems}
\label{sec:exp_hyb_framework}
In this section, we experimentally test the ability of our hybrid model learning framework to detect new contact modes and learn and/reuse them for variable impedance control within that mode. We demonstrated the need for having a hybrid model for learning the dynamics of piecewise-continuous systems in our previous work \citep{sidhik_learning}, where we showed how using a single model is insufficient for accurate trajectory tracking in the presence of such dynamics. Here, we evaluate the ability of the mode-detection component of our framework in identifying known and unknown contact modes and using appropriate forward models for variable impedance control in that mode (\textbf{H2}).

\begin{figure}[tb]
  \begin{center} 
    \includegraphics[width=\columnwidth]{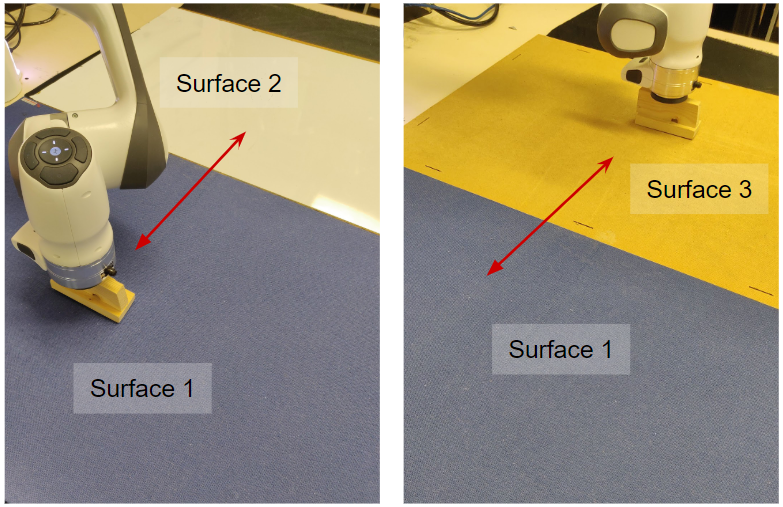}
    \caption{Manipulator sliding an object in a pattern along three surfaces with different friction.}
    \label{fig:example}
  \end{center}
  \vspace{-1em}
\end{figure}

The robot was asked to slide an object along a desired trajectory, and it experienced three previously unseen surfaces with different values of friction (see Figure \ref{fig:example}). We expected the robot to identify a transition to each new mode (i.e., each surface) and incrementally build a dynamics model for the mode while operating under high stiffness. Once the dynamics models for a mode had been built, we expected the robot to respond to subsequent transitions to this mode by using the corresponding dynamics model.  The feature representation we used for clustering to distinguish between the contact modes was $\frac{\lVert F_{ee}^t\rVert}{{R^t}}$ as mentioned in Section \ref{sec:probform-moderecog}, which helps to differentiate between the surfaces by estimating distributions for their friction coefficients.

\begin{figure*}
\centering
\includegraphics[width=\linewidth]{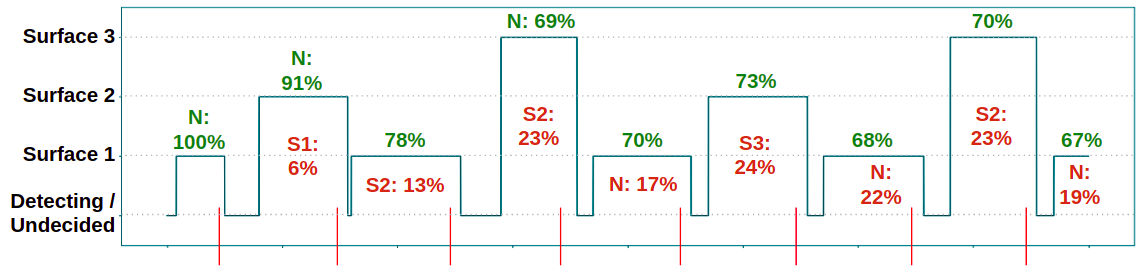}
\caption{Modes detected and their confidence values; red vertical lines on the x-axis indicate actual mode transitions. The number on top of a peak (in green) indicates the confidence with which the transition was identified; the number below a peak (in red) corresponds to the mode with the next highest confidence. ``N'' indicates a transition to a new mode.}
\label{fig:diff_surf_mode_switch}
\end{figure*}

\begin{figure}[tb]
\centering
\includegraphics[width=\columnwidth]{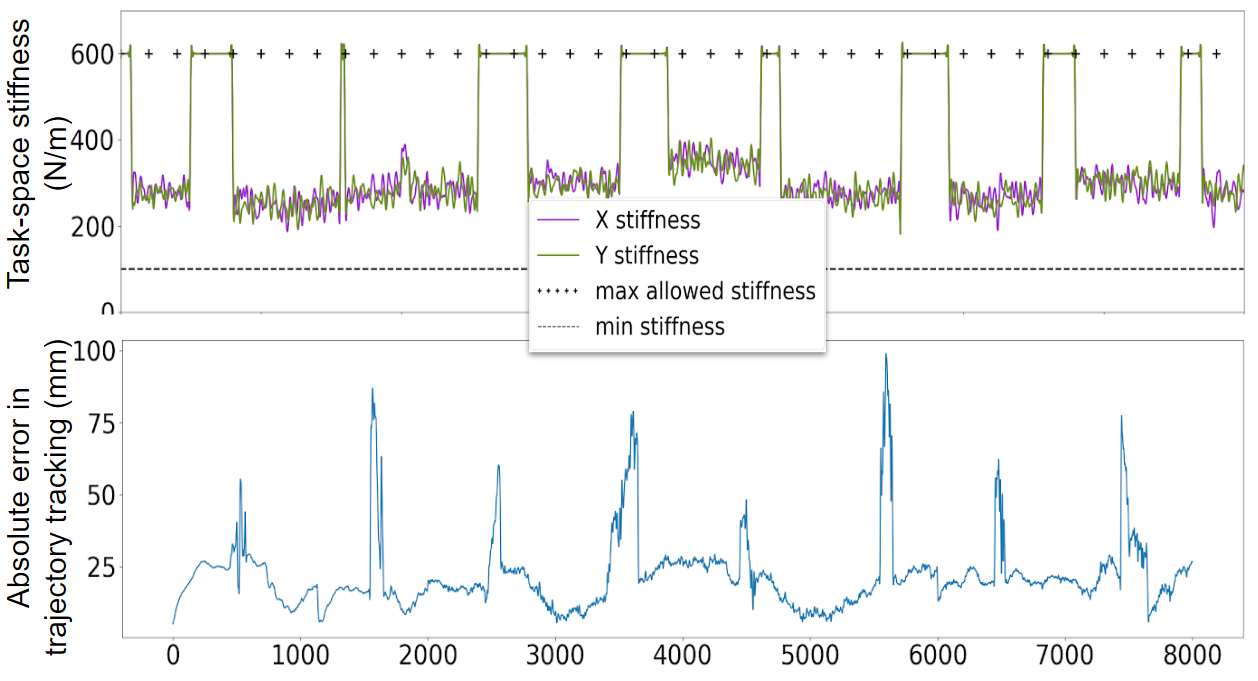}
\caption{Performance of the hybrid mode learning and controller for changing-surface task. \textbf{Top:} controller stiffness. \textbf{Bottom:} absolute error in trajectory tracking. The spikes during trajectory tracking correspond to a temporary, incorrect feed-forward prediction by the previous model after the guard regions. }
\label{fig:diff_surf_traj_stiff}
\end{figure}

Figure~\ref{fig:diff_surf_mode_switch} shows the robot's ability to detect mode changes in one trial of this experiment. The robot was able to identify transitions to existing or new modes with high confidence. In each instance, the second best choice of mode was associated with a much lower value of confidence. The results also show that our hybrid framework and feature representation make performance robust to changes in the direction of motion, i.e., a new mode is not identified when the manipulator moves over a previously seen surface in a new direction. There was some similarity in the confidence values for surfaces 2 and 3 (S2 and S3 in the plot) because of the similarity in their friction values. 

Figure~\ref{fig:diff_surf_traj_stiff} shows the trajectory tracking error and the values of the stiffness parameters of the controller during the trial. The peaks in the trajectory error plot correspond to a sudden change of surface. During each such instance, the predictions made by the dynamics model of the previous mode caused a momentary error in the trajectory tracking ability, until the robot switched to the high-stiffness mode and identified the current mode; the robot then used suitable low(er) stiffness to complete the task. It can also be seen that switching to a previously seen mode requires a much shorter period of high stiffness compared with building a new dynamics model. These results show that the framework is able to identify and learn the interaction dynamics for different contact modes, supporting (\textbf{H2}). In our previous work, we also demonstrated the ability of the framework to be able to identify and learn the contact modes for the robot performing sliding tasks using different types of contacts (e.g. edge contact, surface contact, etc.) while being robust to changes in direction of motion and applied normal force \citep{sidhik_learning}.

\subsection{Need for online learning for hybrid systems} 
\label{sec:exp-hyb-sys}

To explore the need for incremental hybrid models for systems with piece-wise continuous dynamics (\textbf{H3}), we compared our hybrid model learning strategy with a baseline framework that performs offline long-term prediction of dynamics~\citep{khader2020data}. It identifies different dynamic modes in the task from a training dataset of desired motion, uses multi-class Support Vector Machines to build guard function to predict mode changes, builds separate Gaussian Process (GP) dynamics model for each mode, and provides a probabilistic algorithm for multi-step prediction of joint-space state variables for the changing-contact manipulation task. This baseline framework learns a hybrid model from a training set of repeated task demonstrations. It was proposed as a data-efficient way of predicting long-term state evolution of joint-space dynamics for tasks involving discontinuous dynamics; it was \textit{not} designed as a solution for real-time control in such tasks. However, it does present an interesting solution for the core problem of control of tasks with discontinuous dynamics. We use it as a baseline to show the need for real-time control and learning in changing-contact robot manipulation.

We used the PyBullet simulation environment~\citep{coumans2021} to setup a task similar to the one used with the baseline framework. As shown in Figure \ref{fig:pb-shahbaz-task}, the  robot (with a block fixed to the end-effector) approaches a table to make contact with it, slides along the table until it collides with a wall, and slides on the table along the wall. For evaluating \textbf{H3}, we wanted to test the following claims: (i) Unlike our method for learning piecewise-continuous dynamics, the baseline approach is unreliable for tasks with discontinuous dynamics unless the robot uses high-stiffness control or is trained and tested in the same environment; (ii) real-time update of predictive (dynamics) models from our approach provides more reliable performance than the baseline's offline, long-term prediction of dynamics in the presence of unexpected environmental changes.



\subsubsection{Long-term prediction of joint dynamics:}

\begin{figure}[t]
\centering
\begin{subfigure}{\linewidth}
  \centering
  \includegraphics[width=\linewidth]{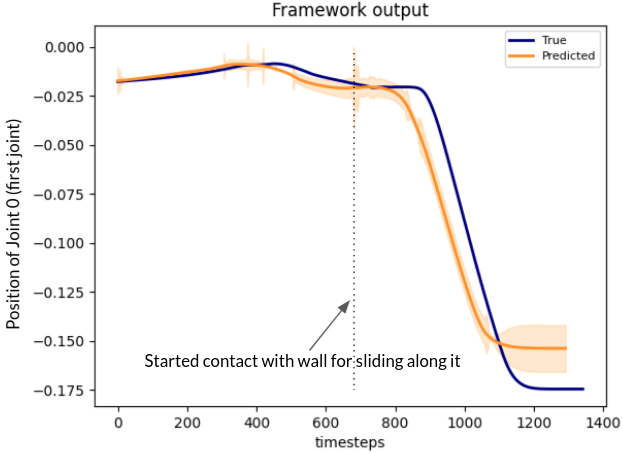}
  \vspace{-1.5em}
  \caption{System trained with wall friction 0.6, tested with wall friction 0.9.}
   \label{fig:shahbaz-js-1}
\end{subfigure}%

\begin{subfigure}{\linewidth}
  \centering
  \includegraphics[width=\linewidth]{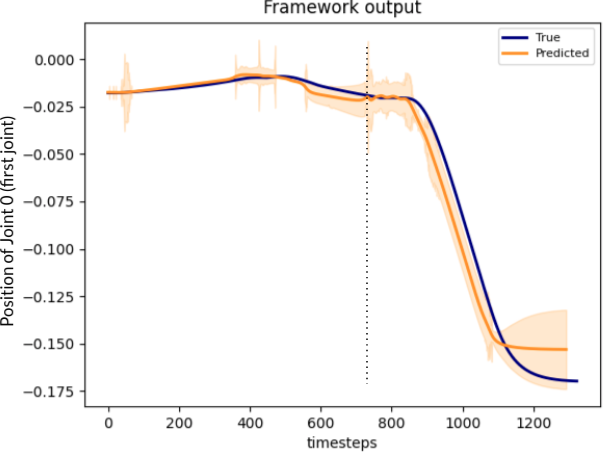}
  \vspace{-1.5em}
  \caption{System trained with wall friction 0.1-0.7, tested on wall friction 0.9.}
  \label{fig:shahbaz-js-2}
\end{subfigure}
\vspace{-0.5em}
\caption{Long term prediction of one joint's position using baseline framework. Dotted vertical line indicates robot's contact with wall obstacle.}
\label{fig:shahbaz-js-1and2}
\end{figure}


We first trained the baseline system with a dataset of trials in which the robot used a fixed medium-stiffness controller and the wall (second collision, red block in Figure \ref{fig:pb-shahbaz-task}) friction was fixed to $0.6$. Testing was then done with an unseen higher friction value (0.9) and the same controller. Figure \ref{fig:shahbaz-js-1} shows the prediction and actual values of the positions of one of the relevant joints in the task (first joint). Being unaware of the change in friction, the offline framework predicted the joints to move as freely as it did during training. The true joint positions were, however, affected by the higher friction resulting in the ``true" values lagging behind the ``predicted" values.

\begin{figure}[tb]
  \centering
  \includegraphics[width=\linewidth]{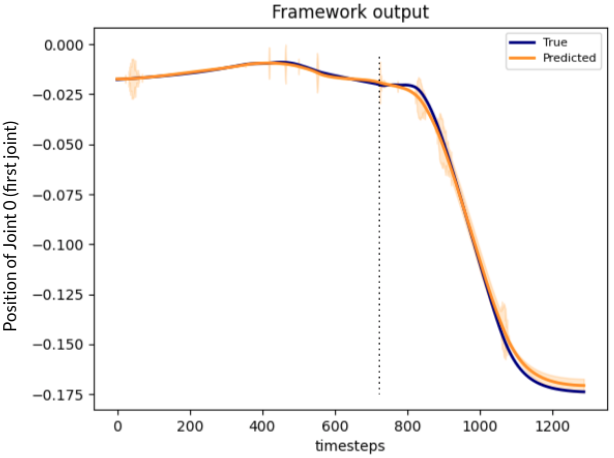}
  \vspace{-1em}
  \captionof{figure}{Long term prediction of the position of one joint produced by the baseline framework. System trained and tested using a constant high-stiffness controller (friction values as in Figure \ref{fig:shahbaz-js-2}).}
    \label{fig:shahbaz-js-3}
\end{figure}

For further exploration, the training set was modified to include wall friction values between $0.1-0.7$ at increments of $0.1$; the system was then tested on a wall with friction 0.9. This results in a prediction that is similar to the previous set of experiments, but with a wider band of uncertainty around the prediction due to the larger variability in the training data (Figure \ref{fig:shahbaz-js-2}). In addition, when a high stiffness controller is used for training and testing, the robot follows the trajectory quite closely regardless of friction. This results in lower variability across trials (marked by the narrower uncertainty band around the prediction), and more accurate predictions (Figure \ref{fig:shahbaz-js-3}). These results indicate that learning dynamics in the space of joint positions and velocities poses a high-dimensional problem; the learned models do not capture the interaction dynamics accurately, and predictions are reliable only when used with a high stiffness controller.



\begin{figure}[tb]
  \centering
  \includegraphics[width=\linewidth]{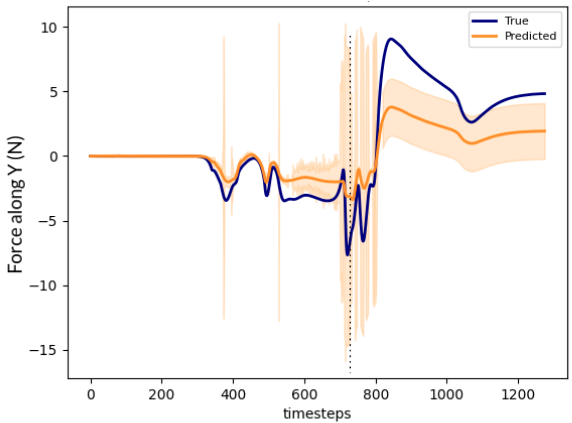}
  \vspace{-1em}
  \captionof{figure}{Long term prediction of end-effector force along Y-axis (along the wall). System  trained with wall friction values between $0.1-0.7$, tested on wall with friction 0.9.}
    \label{fig:shahbaz-ts-lt}
\end{figure}



The main reason for the baseline framework's poorer performance in the presence of unseen environments is the fact that the system is unaware of the environment's dynamics before it performs the prediction. This is not a flaw of the framework as it was not designed for such use cases. 
However, the experiments show that for accurate long-term prediction of joint-space dynamics, either the environment has to remain unchanged and/or the robot has to use a high-stiffness controller. For environments whose dynamics can change in previously unseen ways, the system will at least have to be trained with many examples of different environments. Our framework addresses these issues by learning the the dynamics in the end-effector space (i.e., task space instead of joint space) and has access to a more direct measurement of the interaction dynamics in the form of end-effector forces and torques.


\subsubsection{Long-term prediction in the task-space:}
Next, the baseline framework was modified to consider interaction dynamics in the task space by reformulating the framework's GPs (that model each mode's dynamics) to model $p(F_{ee_{t+1}} | F_{ee_{t}}, \dot{x}_{t}])$, similar to the forward model of our base control framework (Section \ref{sec:probform-lowlevel-model}). 

The system was again trained with examples of the state evolution when the surface friction was between 0.1 and 0.7, and then was compared with the measurements taken when the surface friction was 0.9. Figure \ref{fig:shahbaz-ts-lt} shows the measured and predicted force measurements along the Y-axis (the direction along which the wall's friction acts). The spikes in Figure \ref{fig:shahbaz-ts-lt} at $\approx 400$ along the X-axis corresponded to when the robot made contact with the table, and the spikes between $700-800$ corresponded to when the robot collided with the wall. We observed that the actual values of friction force after contact with the wall did not fall within the predicted band of uncertainty. Also, the effect of different levels of friction was more observable in the task space, indicating that it is more meaningful to represent the interaction dynamics for changing-contact tasks in the task space. This experiment suggests that interaction dynamics cannot be predicted reliably without some form of real-time feedback. 

\begin{figure}[tb]
\centering
\begin{subfigure}{\linewidth}
  \centering
  \includegraphics[width=\linewidth]{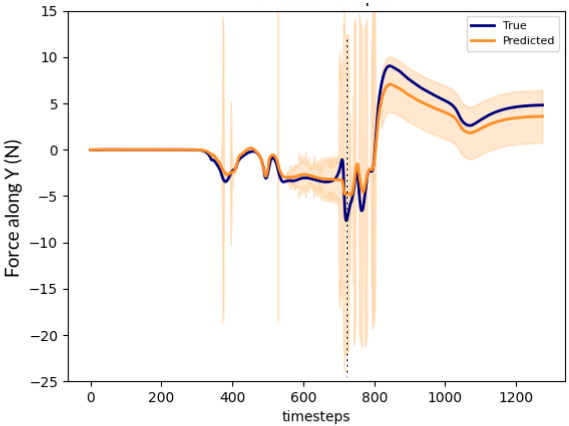}
  \caption{Modified baseline framework's prediction.}
   \label{fig:shahbaz-ts-st}
\end{subfigure}%

\begin{subfigure}{\linewidth}
  \centering
  \includegraphics[width=\linewidth]{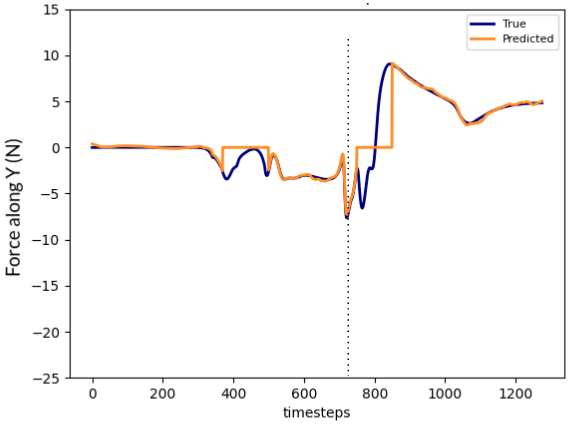}
  \caption{Predictions from the forward model of our base controller.}
  \label{fig:shahbaz-hsavic-ts}
\end{subfigure}
\vspace{-0.5em}
\caption{Prediction of end-effector force along Y-axis (along the wall). Each system was trained with wall friction values between $0.1-0.7$, and tested on a wall with friction 0.9. Dotted vertical line indicates when the robot made contact with the wall (obstacle).}
\label{fig:shahbaz-ts-st-compare}
\end{figure}



\subsubsection{One step prediction in task-space using real-time feedback:}

In the next experiment, real-time measurements from the robot ($[F_{ee_{t}}, \dot{x}_{t}]$) were provided as feedback to the baseline framework during testing. At each timestep, the system then only had to predict sensor values for the next timestep using the dynamics model for the identified mode. Although this strategy produced better results (see Figure \ref{fig:shahbaz-ts-st}) by extrapolating from the learned model to handle previously unseen values, the predictions are still not accurate. This is because the baseline framework's model is not incrementally updated to account for the new environment dynamics. 
Our hybrid framework, on the other hand, supports incremental, real-time updates to the dynamics models. This resulted in more reliable predictions for the same experiment (see Figure \ref{fig:shahbaz-hsavic-ts}) and thus more accurate trajectory tracking in the presence of discretely changing dynamics and previously unseen environments.

Note that the training of the baseline framework required significant time depending on the density and the length of the data in the training set. The implementation of the GP in this framework used $\approx 55$ minutes training time on an 8-core, 16GB RAM computer without using a GPU, for a dataset consisting of 40 trials of the task mentioned above; the long-term prediction process during testing takes $\approx 20$ minutes.

Overall, the experiments discussed above demonstrate that the incremental, real-time revision of the predictive (dynamics) models is critical for changing-contact manipulation tasks, particularly when the learning is done in the task space and suitable feedback of the system state is available (and used). The results of these experiment support hypothesis \textbf{H3}. The observed results are also influenced by the choice of the state space and feature representation used for control and learning. A hybrid control framework that supports such control and learning is better able to adapt to previously unseen variations in the dynamics and provide smooth control of changing-contact robot manipulation tasks.

\subsection{Evaluating framework capability in producing smooth overall motion}
\label{sec:exp-smooth_motion}

In this section, we test the ability of the overall framework (AVIC + hybrid model learning + contact-changing-handling) to anticipate mode transitions and produce smooth overall motion by switching to appropriate transition phase controllers in the anticipated regions (\textbf{H4}). We demonstrated the ability of the contact-change-handling module in incrementally improving the estimates of contact location and learning appropriate approach velocities for impacts in our previous work \cite{sidhik:iros21}.

\begin{figure}[ht]
\centering
\includegraphics[width=0.9\linewidth]{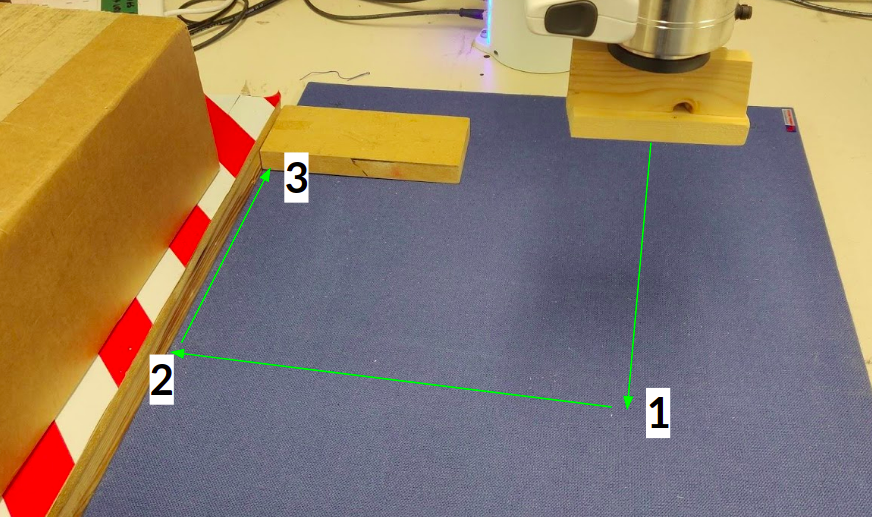}
\caption{A sliding task that involves making contacts with the table's surface at "1", with a wall at "2", and with another object at "3".}
\label{fig:collision_task_setup}
\end{figure}

\subsubsection{Framework in the presence of multiple collisions:}
To evaluate the overall framework and the resulting dynamics on a physical robot performing tasks involving multiple collisions, the robot (with a wooden block attached to end-effector) was asked to move vertically down to the table (contact 1), slide along y-axis (the table's surface) to a wall (contact 2), and slide along the wall (while in contact with  the table's surface) to another obstacle (contact 3), as shown in Figure \ref{fig:collision_task_setup}. The robot was provided significantly incorrect initial guesses of the contact positions with substantial noise (see Table \ref{tab:contact_preds}). The robot had to repeat the task while reducing the deviation from the given motion pattern by improving its estimate of the contact positions. The robot also had to modify its approach velocity from the initial value of $0.05\,m/s$ to produce a desired impact force of $8\,$N. Since each contact in the task is in the presence of different environment dynamics (e.g., motion in free space, motion against surface friction), the velocity required to attain the desired impact force was expected to be different. The robot also had to incrementally update its approach velocity for each contact using gradient descent till the desired velocity for that environment was achieved. Furthermore, the robot had to perform all the trials with smooth overall motion dynamics with minimum spikes in the velocity or acceleration profiles.

\begin{figure*}[!th]
  \begin{center}
    \begin{subfigure}{0.9\textwidth}
      \includegraphics[width=0.9\textwidth]{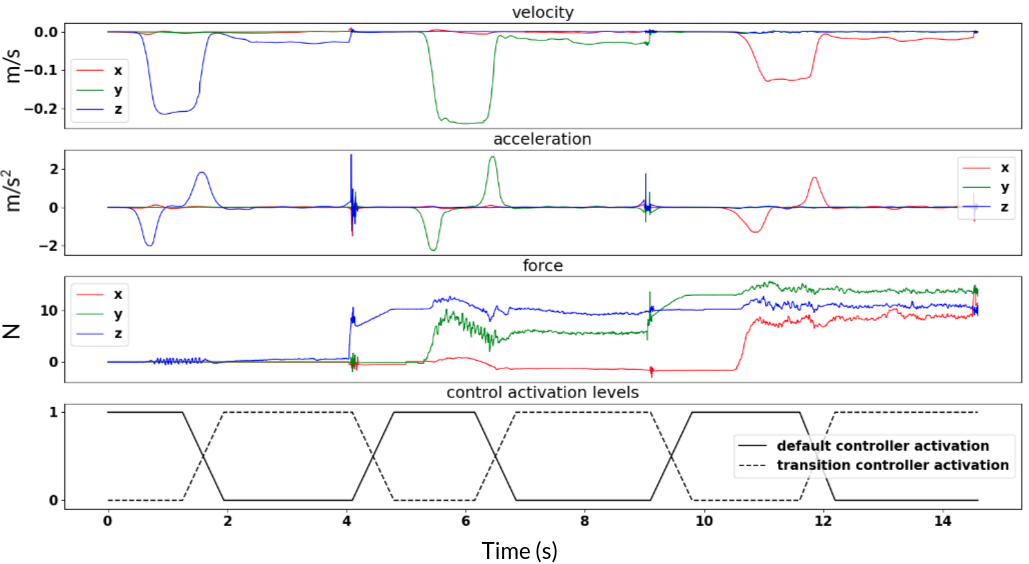}
      \vspace{-1.5em}
      \caption{Experiment trial 1.}
      \label{fig:expt_first_trial}
    \end{subfigure}
    \begin{subfigure}{0.9\textwidth}
      \includegraphics[width=0.9\textwidth]{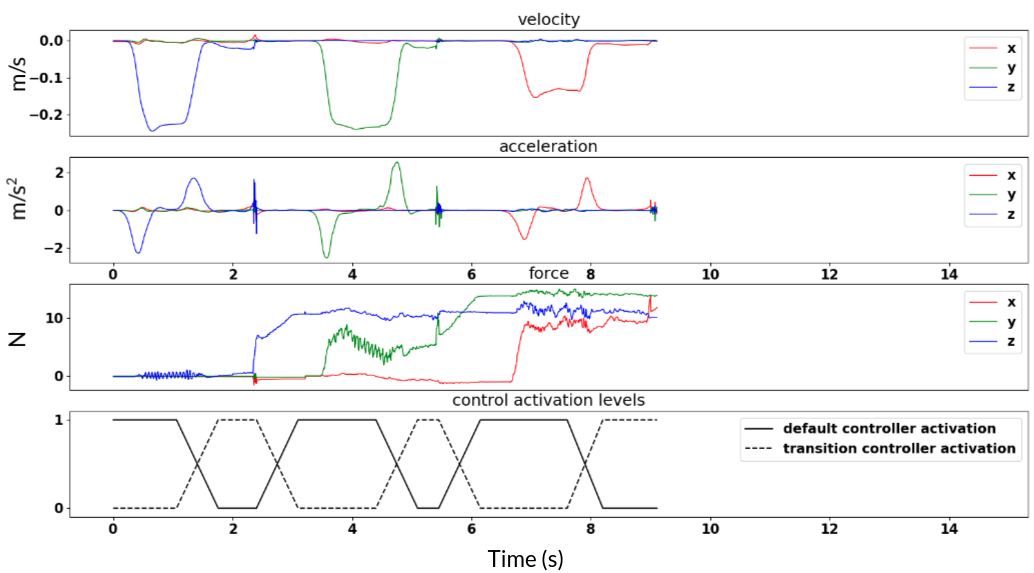}
      \vspace{-1.5em}
      \caption{Experiment trial 5.}
      \label{fig:expt_last_trial}
   \end{subfigure}
  \end{center}
  \vspace{-1em}
  \caption{Velocity, acceleration, force, and controller activation levels in: (a) experimental trial 1; (b) experimental trial 5. Use of our framework reduces uncertainty in estimates of contact positions, reduces the time spent using the transition-phase controller, and reduces discontinuities.}
  \label{fig:expt_all_collisions}
\end{figure*}

Figure \ref{fig:expt_first_trial} shows the velocity, acceleration, and EE force in the first trial, and Figure \ref{fig:expt_last_trial} shows these values after five trials. The results in these figures and in Table \ref{tab:contact_preds} show that the uncertainty in the estimate of the contact positions is reduced, as indicated by a significant reduction in the size of the covariance ellipsoids, and the robot spends significantly less time using the transition-phase controller and the associated lower velocity. The last plot in Figure \ref{fig:expt_first_trial} and Figure \ref{fig:expt_last_trial} show the activation of the default controller and the transition-phase controller. The overall task was completed in $9.2~s$ in the fifth trial as opposed to $14.4~s$ in the first trial. These results support \textbf{H4}.

\begin{table}[tb]
\centering
\begin{tabular}{lcr}
\toprule
 Prediction Error (m) & Initial & Final (trial 5)  \\
\midrule
Contact 1 (Z-axis) & 0.12 $\pm$ 0.3  & 0.016 $\pm$ 0.039 \\
Contact 2 (Y-axis) & 0.09 $\pm$ 0.2 & 0.011 $\pm$ 0.04 \\
Contact 3 (X-axis) & 0.1 $\pm$ 0.2 & 0.018 $\pm$ 0.036 \\
\bottomrule
\end{tabular}
\caption{Error in the estimated contact location along the most significant axis for the contact (in parenthesis) in the first and fifth trials of the task in Figure \ref{fig:collision_task_setup}. The value along the diagonal of the corresponding covariance matrix is shown as the  standard deviation ($\pm$ term).}
\label{tab:contact_preds}
\end{table}
  
 The covariance ellipsoids converged in the first three trials of the task, but the task was repeated to evaluate the ability to compute and set the approach velocity for different transition-phase controllers. The robot converged to a suitable approach velocity for the first contact (from motion in free space) in five iterations. It was, however, difficult for the robot to adjust its approach velocities for contacts 2 and 3, which required the robot to use force control along one and two directions (respectively). Contact 3 was particularly challenging because it involved sliding along two different surfaces, resulting in very noisy readings from the force-torque sensor due to the different values of frictional resistance offered by the two surfaces. Since the impact force was along the same direction as friction, it was more difficult to isolate the impact force from the force due to surface friction, which made updating the approach velocity more challenging.

\subsubsection{Framework effectiveness for impact-less transitions:}

\begin{figure*}[!tb]
\centering
\includegraphics[width=0.7\linewidth]{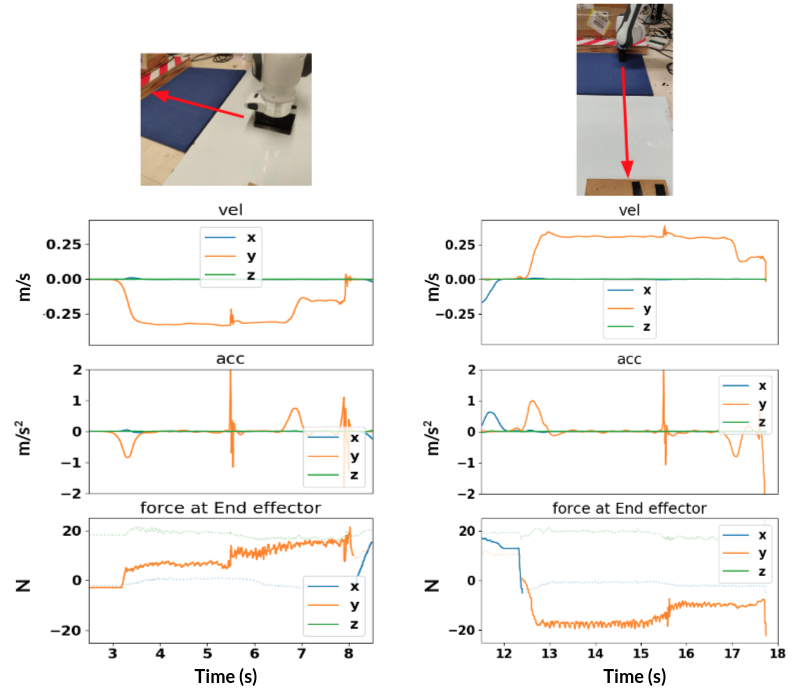}
\caption{Selected sections of the velocity (top), acceleration (middle), and force (bottom) curves in the first trial of the task; the robot is unaware of the impact-less transitions in the task beforehand. \textbf{Left:} Curves for the first impact-less transition (motion `2' in Figure \ref{fig:exp-impactless_task}); \textbf{Right:} Curves for the second impact-less transition (motion `4' in Figure \ref{fig:exp-impactless_task})}
\label{fig:expt_cc_task2_trial_1}
\end{figure*}
In this experiment, the objective was to test the effectiveness of the framework in learning to predict contact changes that are not due to collisions, but due to impact-less transitions.

The experiment involved the robot performing a similar changing-contact experiment as before, but this time, the surface along which the robot has to slide suddenly changes without the knowledge of the robot (see Figure \ref{fig:exp-impactless_task}). The location of the surface switches are not known beforehand, and the robot has to learn to anticipate them in subsequent trials once it realizes that there are previously unknown contact changes in the provided plan. The task involved the robot approaching a surface (surface A) from top, and sliding along the surface to make contact with a wall as before; however, the surface changes midway to one with higher friction (surface B) unknown to the robot. The robot then has to slide along the wall, and then move to another wall parallel to it, on the way to which the robot will slide surface B to A again. As in the previous experiment, the robot is provided with initial guesses of where \textit{collisions} occur in the task, but it is not provided any knowledge about the changes in surfaces; this, the robot has to detect in its first trial and should then be able to predict and handle in subsequent trials.

The desired behavior from the robot is that it learns to anticipate collisions as well as impact-less contact changes, and smoothly switches to an appropriate controller, that reduces the discontinuities in the motion dynamics during task execution.

\begin{figure*}[!th]
  \begin{center}
    \begin{subfigure}{0.85\textwidth}
      \includegraphics[width=\textwidth]{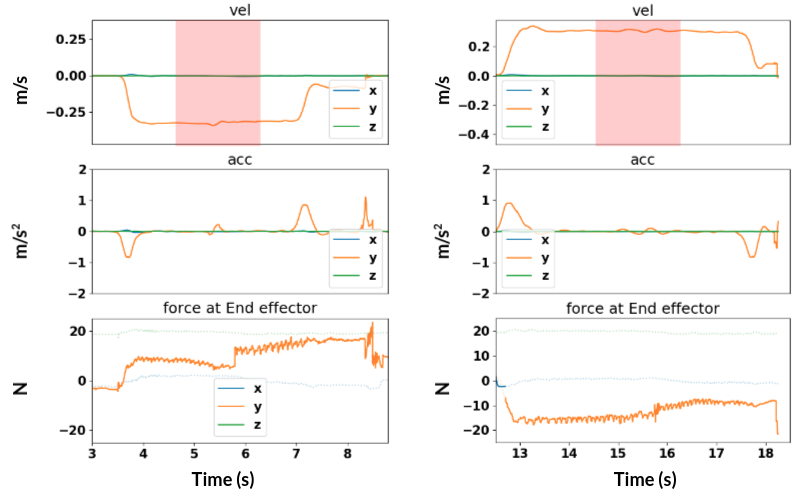}
      \vspace{-1.5em}
      \caption{Experiment trial 2.}
      \label{fig:expt_cc_task2_trial_2}
    \end{subfigure}
    \begin{subfigure}{0.8\textwidth}
      \includegraphics[width=\textwidth]{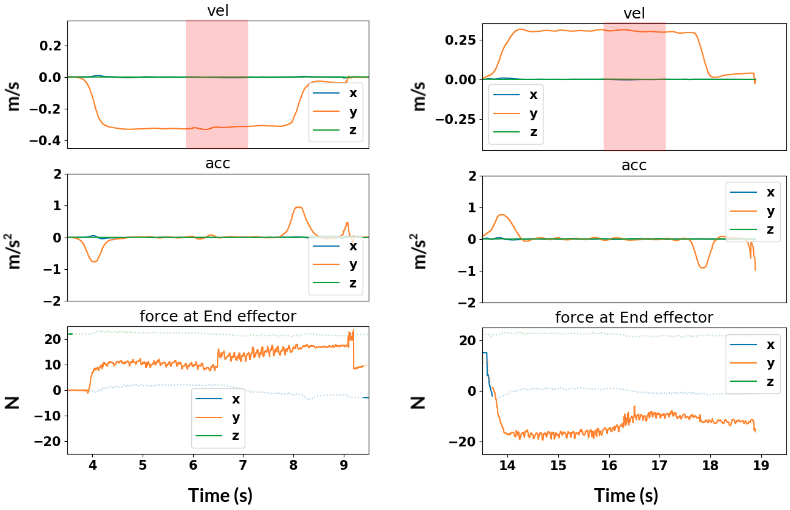}
      \vspace{-1.5em}
      \caption{Experiment trial 3.}
      \label{fig:expt_cc_task2_trial_3}
   \end{subfigure}
  \end{center}
  \vspace{-1em}
  \caption{Selected sections of the velocity (top), acceleration (middle), and force (bottom) curves in trials 2 and 3. \textbf{Left:} Curves for the first impact-less transition (motion `2' in Figure \ref{fig:exp-impactless_task}); \textbf{Right:} Curves for the second impact-less transition (motion `4' in Figure \ref{fig:exp-impactless_task})}
  \label{fig:expt_cc_task2_trials_2and3}
\end{figure*}

In this experiment, we focus on the performance of the framework in the regions where the surfaces change, i.e., where there are impact-less mode transitions. This occurs at two sections in a trial of the task:  Figure \ref{fig:expt_cc_task2_trial_1} shows the velocity, acceleration, and end-effector force profiles in these regions measured during the first trial of the task. As expected, the robot experiences sudden spikes in velocity and acceleration (jerk) when the surface changes unexpectedly. Recall from previous chapter that this is due to the wrong predictions provided by the dynamics model which provides a feed-forward term for the controller that either overestimates or underestimates the environment forces in the new mode. When the robot detects such a discontinuity, it identifies this as a contact change and quickly switches to a high-stiffness controller to identify the new mode it is in. Seeing that this is a new surface which it hasn't seen before, the robot learns a new dynamics model for the mode which it uses as a forward model for the base controller in that mode. The surface of the table changes twice during the task; however, for the second surface change, the robot recognises the new dynamics mode as a surface that it has seen before (surface A).

In the next trial (Figure \ref{fig:expt_cc_task2_trial_2}), the robot expects these mode switches  (the anticipated region is marked in the figure), and because it anticipates an impact-less contact change, the robot switches to a high stiffness controller which would help the robot to identify the new mode quickly while also reducing sudden velocity-acceleration spikes during motion. As desired, the size of the anticipated region reduces in subsequent trials (Figure \ref{fig:expt_cc_task2_trial_3}). These results further support hypothesis \textbf{H4}.

An interesting phenomenon that was observed during the experiment was that the robot using this \textit{high-stiffness controller} is able to easily detect that a contact change has occurred when it slides from a surface of lower friction (A) to that of higher friction (B); however, it does not always detect that a mode change has occurred when sliding from B to A. This was understood to be because of the fact that when sliding a block from a surface of low friction to that of high friction, the sudden increase in friction at the region of the block in contact with the new surface offers the highest resistance, and contributes to the frictional resistance of the block. This change is sudden and pronounced (as seen in figure). On the other hand, when sliding from a high-friction surface to a low-friction surface, the trailing part of the block is in contact with the high friction surface which still offers resistance. This causes the robot to feel that the frictional resistance is being reduced gradually and not suddenly as in the other case. This lack of discontinuity in the sensed forces, and the absence of spikes in velocity due to the high-stiffness controller motivated the need to add another component to the contact-change-detection module of the framework: a threshold for the error between the measured end-effector force and the prediction from a (non-updating) forward model for that mode.


\section{Conclusion} 
\label{sec:conclusion}
Many robot manipulation (and human manipulation) tasks are changing-contact manipulation tasks. They are characterized by piece-wise continuous interaction dynamics, with discontinuities due to changes is contacts, surfaces, and other factors, and continuous elsewhere. While it is possible to construct a hybrid framework with continuous dynamics within each of a set of discrete modes, it is difficult to provide comprehensive information about the modes in practical tasks.

In this paper, we presented an adaptive, hybrid framework for changing-contact manipulation tasks inspired by studies in human motor control. The framework has a high-level mode detection module which can recognize existing modes and identify new contact modes. Each such mode is characterized by a suitable predictive/forward dynamics model, control law, and relevance condition. The forward model is learned and revised incrementally during run-time, with the error between the predictive and actual measurements (of end-effector forces and torques) is used to adapt the gain parameters in the control law. Our framework also includes a contact handling module that uses a Kalman filter to incrementally improve estimates of contact positions. These contact position estimates help revise the velocity profile to minimize the time spent in a transition-phase controller, ensure smooth transition to and from this controller, and achieve a desired force on impact. Our representational choices enable us to simplify and address the associated challenges reliably and efficiently.

We showed the effectiveness of the overall framework in the context of a physical robot performing manipulation tasks that involve multiple contact changes. We have experimentally demonstrated the need for adaptive control and learning strategy in the presence of tasks and environments with a mix of continuous and discontinuous dynamics. We compared our framework's performance with a representative method from each of three different classes of adaptive control methods in the existing literature, and with a sophisticated framework for offline, long-term prediction in the context of discontinuous dynamics, to highlight the capabilities of our framework. We did so in simulated as well as physical systems where a robot is executing motion patterns that involve making and breaking contacts on a tabletop. Results from our experiments support our claim that a hybrid framework that supports adaptive control and learning enables smooth control of changing-contact robot manipulation tasks.

\bibliographystyle{SageH}
\bibliography{references}






\end{document}